\pdfoutput=1
\documentclass[sigconf,nonacm]{acmart}
\usepackage{graphicx} 
\graphicspath{{figures/}}
\usepackage{booktabs} 
\usepackage{array} 
\usepackage{caption} 
\usepackage[title]{appendix}
\usepackage{algorithm}
\usepackage{algpseudocode}
\usepackage{amsmath}
\usepackage{amsfonts}
\usepackage{multirow}
\usepackage{tablefootnote}
\usepackage{wrapfig}
\usepackage{arydshln}
\usepackage{url}
\usepackage{subcaption}
\usepackage{float}
\begin{document}
\fancyhead{}
\pagestyle{empty}
\title{ScarceGAN: Discriminative Classification Framework for Rare Class Identification for Longitudinal Data with Weak Prior}
\author{Surajit Chakrabarty}
\email{surajit.chakrabarty@games24x7.com}
\affiliation{%
  \institution{\textit{Artificial Intelligence and Data Science, Games24x7}}
  \country{India}
  }
  
\author{Rukma Talwadker}
\email{rukma.talwadker@games24x7.com}
\affiliation{%
  \institution{\textit{Artificial Intelligence and Data Science, Games24x7}}
  \country{India}
  }
  
\author{Tridib Mukherjee}
\email{tridib.mukherjee@games24x7.com}
\affiliation{%
  \institution{\textit{Artificial Intelligence and Data Science, Games24x7}}
  \country{India}
  }
\begin{abstract}
This paper introduces ScarceGAN which focuses on identification of extremely rare or scarce samples from multi-dimensional longitudinal telemetry data with small and weak label prior. We specifically address: (i) severe scarcity in positive class, stemming from both underlying organic skew in the data, as well as extremely limited labels; (ii) multi-class nature of the negative samples, with uneven density distributions and partially overlapping feature distributions; and (iii) massively unlabelled data leading to tiny and weak prior on both positive and negative classes, and possibility of unseen or unknown behavior in the unlabelled set, especially in the negative class. Although related to PU learning problems, we contend that knowledge (or lack of it) on the negative class can be leveraged to learn the compliment of it (i.e., the positive class) better in a semi-supervised manner. To this effect, ScarceGAN re-formulates semi-supervised GAN by accommodating weakly labelled multi-class negative samples and the available positive samples. It relaxes the supervised discriminator's constraint on exact differentiation between negative samples by introducing a `leeway' term for samples with noisy prior. We propose modifications to the cost objectives of discriminator, in supervised and unsupervised path as well as that of the generator. For identifying risky players in skill gaming, this formulation in whole gives us a recall of over 85\% ($\sim$60\% jump over vanilla semi-supervised GAN) on our scarce class with very minimal verbosity in the unknown space. Further ScarceGAN outperforms the recall benchmarks established by recent GAN based specialized  models for the positive imbalanced class identification and establishes a new benchmark in identifying one of rare attack classes (0.09\%) in the intrusion dataset from the KDDCUP99 challenge. We establish ScarceGAN \footnote{This is the PRE-PRINT version of the work accepted for publication in the  ACM CIKM'21. The final version is published by ACM and is available at:
https://doi.org/10.1145/3459637.3482474} to be one of new competitive benchmark frameworks in the rare class identification for longitudinal telemetry data.
\end{abstract}

\maketitle

\section{Introduction}
\label{sec:intro}
Learning non-linear classifiers from unlabelled or only partially labelled data is a long standing problem in machine learning. The premise behind bootstrapping a model from a partially labelled data and then incremental learning from the unlabelled data is to further leverage information present in the unlabelled set to refine the class structures and obtain robust class decision boundaries. Most of the work in computer vision, only to name a few \cite{rumigan,catgan,genpu} have exploited this premise very well. However, in images, the model learnt structural separability can be very well scrutinized visually. A car looks different from a chair! 
However, 
for telemetry data, which is also time varying, 
the level of complexity is different. 
What separates a positive data sample from negative is 
defined by how the data has evolved over time. 

The problem gets exacerbated further with scarcity of positive class samples, because of organic skewness in the underlying data, as well as label noise. For instance, network intrusion may have evolved from a small unnoticeable event and progressed into a severe but very scarce threat later in time~\cite{intrusion-intro}. 
Similarly, 
in the domain of online skill gaming, consistent and sustained overindulgence at an extreme level may become risky for the players~\cite{gaming}. A player needs to be observed for a sustained time interval on various psycho-physical dimensions: time, money and game play desperation etc., to notice a shift in the play pattern. Although occurrences of extreme risky events are extremely scarce (with $0.05$ - $0.1$\% probability), it is important to identify them in time to ensure and encourage responsible game play with proper intervention.  \textit{This paper focuses on identification of extremely rare or scarce samples from multi-dimensional longitudinal telemetry data with small and weak label prior.}   

\begin{figure}
\begin{minipage}{0.48\linewidth}
\includegraphics[width=\linewidth]{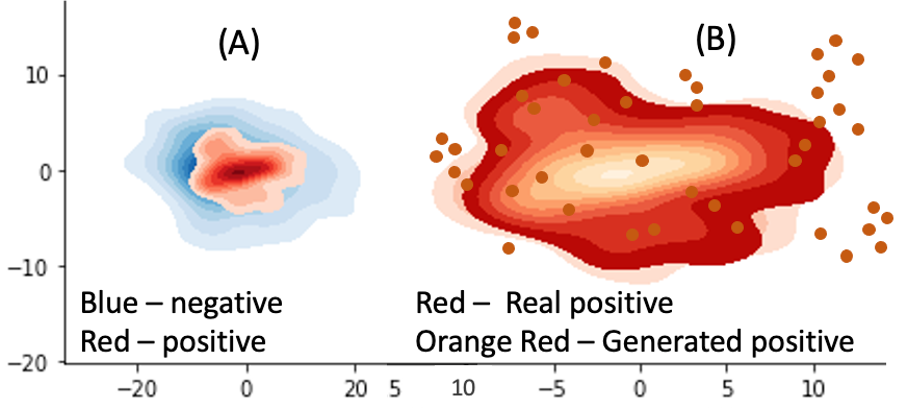}
\caption{(A) no separability between positive-negative samples. (B) poorly augmented positive samples with limited positive data}
\label{kde-pn}
\end{minipage}
\hfill
\begin{minipage}{0.48\linewidth}
\includegraphics[width=\linewidth]{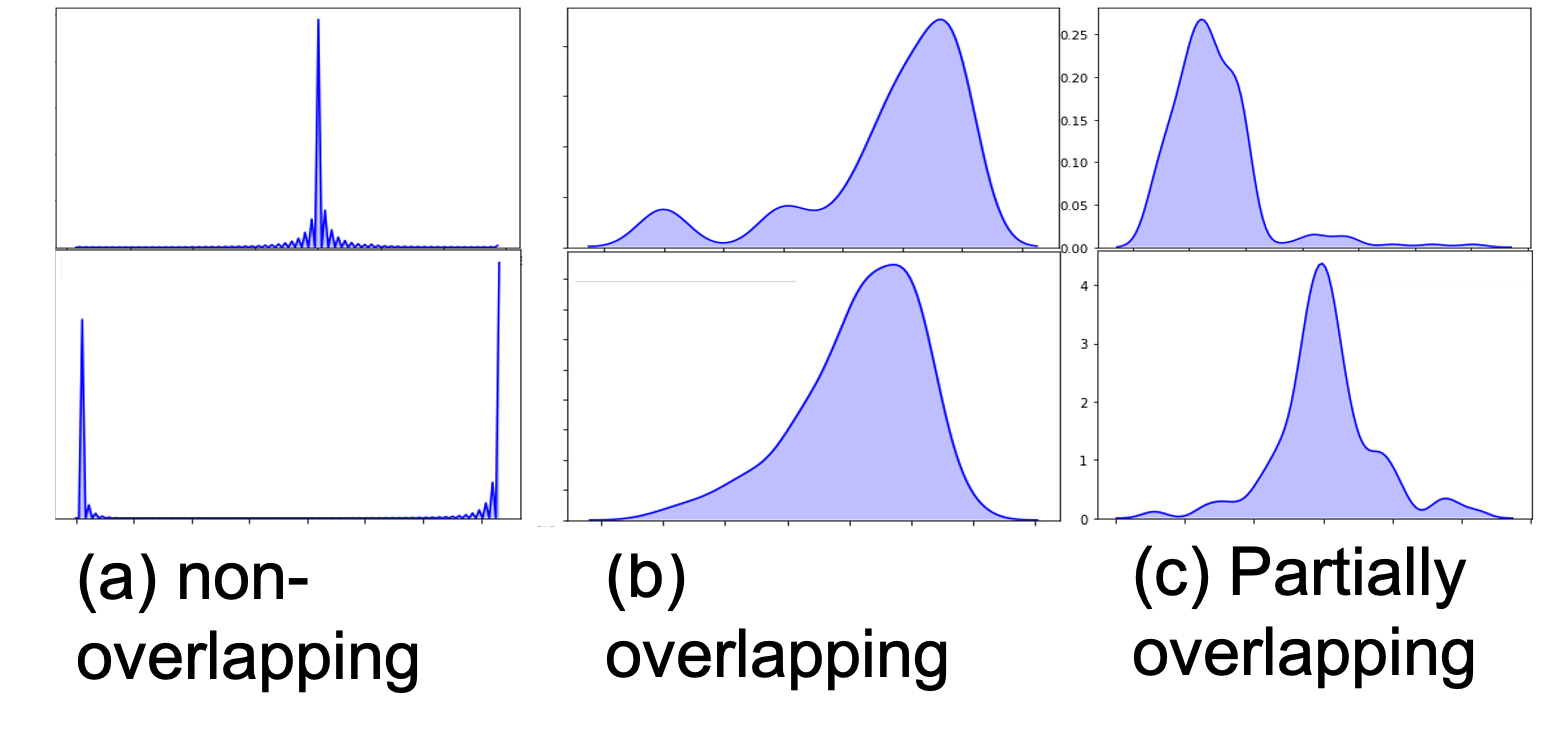}
\caption{Mix of Overlapping and Non-Overlapping  patterns in negative class samples}
\label{fig:mix}
\end{minipage}
\end{figure}

This problem does have analogy from other outlier/risky pattern finding domains like fraudulent transactions \cite{kaggle}, network intrusion \cite{icdm,intrusion2}, digital or social media addiction \cite{social,netflix}, gambling \cite{gambling1,gambling2}, addictive e-commerce shopping \cite{shopaholic} etc. As a working example domain, in the rest of the paper, we will use the risky player identification problem, although our proposition is more generic as we demonstrate in our evaluation in multiple domains. Below, we highlight the specific challenges  addressed in conjunction.
\\
\noindent \textbullet\ \textbf{Scarce positive samples:} We are assigned with the task with a severe class imbalance. This makes it difficult to learn the distribution of the positive prior without avoiding over-fitting. Figure \ref{kde-pn} (B) shows distribution boundaries of the augmented data (light orange) over the training positive samples (dark red) and the relative positioning of the held out positives (as orange circles). Oversampling was done using techniques suggested in \cite{vae-oversampling, smote}. We notice two aspects: 1) no qualitative benefit; 2) newer samples redefine the distribution space, indicating its inherent sparsity and incompleteness. Hence, the focus can be shifted from ``understanding the positive class'' to ``what belongs to the negative class''. 


\noindent\textbullet\ \textbf{Poor inter-class separability:}  Figure \ref{kde-pn}(A) shows the 2D plot of the PCA features obtained on a subset of verified positive and known negative class samples. PCA was performed on features which were identified as the independent predictors of the two classes. Features were aggregated (max) over their  longitudinal counter data. We indeed see that the positive class is a small subset of the negative, in the two dimensional space. This motivates the need for ``capturing time varying aspects of the longitudinal data into the features to bring in the separability''. 

\noindent\textbullet\ \textbf{Mix of Overlapping and non-overlapping  patterns in negative class samples:} Though it seems like a positive-negative class separation problem, we discover that the negative samples are indeed multi-class. Figure \ref{fig:mix}(b) shows that the distributions of the two identified negative classes on a particular independent counter are overlapping and in Figure \ref{fig:mix}(a) the distributions of the same classes on another independent counter are non-overlapping. In the Figure \ref{fig:mix}(c) the distributions on the third counter are partially overlapping. We lack labelled samples of these classes to train a supervised negative multi-class model. We learn that we cannot formulate this as a binary (positive/negative) class problem. A good discriminator ``should learn to separate between the various negative classes to identify the rare positives''.

While individually these challenges may seem novel at a cursory glance, a careful study of the existing literature (Section \ref{sec:related}) highlights a need to establish a benchmark framework to address all these challenges together in the specialized problem space of extremely rare class identification with only a small prior. We propose \textbf{ScarceGAN}, a \textit{unique} framework to \textit{leverage semi-supervised discriminative classifier for PU task from a noisy, weak and tiny amount of negative and positive labelled samples and abundance of unlabelled samples}. The \textbf{key contributions} are described below.

\noindent\textbullet\ \textbf{Increase Scarce Visibility via Divided Focus:} We slice negatives into 3 classes. This reduces the imbalance (e..g., from 0.9:0.1 to 0.3(x3):0.1 for risky player identification problem in skill gaming) and hence improves the ``visibility'' of the positive class. This achieves two objectives. First, it helps the discriminator to  build multiple decision boundaries for robust understanding of the negative classes. Secondly, since positive class is scarce, it now focuses on complementary learning (of the varied negative classes) instead of learning the distribution of the positive class. Via semi-supervised learning, the framework also leverages the structure present in unlabelled samples to refine its feature maps.

\noindent\textbullet\ \textbf{Label-Noise Accommodative Discriminator :} Existing work that assumes noisy inputs, accommodates it so by correcting the labels with supervised data over time \cite{noisylabel,ccgan}. We define a \textit{novel} concept of a `Leeway' term which prevents the confused discriminator from assigning a not so confident negative sample to a positive sample space. This process also generates \textit{complementary} labels for samples with unknown labels for future analysis 

\noindent\textbullet\ \textbf{Pattern Preserving Longitudinal Aggregations:} We bring in the aspect of modeling the longitudinal data using hyper-parameters of a time series model. In doing so, we extract custom parameters from the model's time series curve fitting process summary. This provides us high accuracy on understanding the deviation of a counter pattern from its own normal while adding to the domain specific novelty.

ScarceGAN reformulates Semi-supervised GAN to bring in the aforementioned contributions via introducing new cost objectives of discriminator, in supervised and unsupervised path as well as that of the generator. This formulation as a whole gives us a recall of over 85\% on our scarce class  with very minimal verbosity in the unknown space. Further ScarceGAN outperforms the recall benchmarks on recently published specialized GAN models for identifying imbalanced class and also one of the highly rare attack classes (0.09\%) in the intrusion dataset from the KDDCUP99 challenge. Our source code is available at GitHub \cite{github}.  The rest of the paper is organized as follows: We first review the existing literature in Section \ref{sec:related}. We introduce our new formulation and then derive the ScarceGAN objectives as a direct extension of the semi-supervised GAN framework in the Section \ref{sec:scarcegan}. We present our technique for the pattern preserving longitudinal aggregations in Section \ref{sec:method} followed by a detailed discussion of results.

\section{Related Work}
\label{sec:related}


\begin{table*}[t]
  \centering
  \footnotesize
  \begin{tabular}{|l|l|l|}
\hline
\textbf{Existing Work} & \textbf{Contributions} & \textbf{Shortcomings}\\
\hline
RUMIGAN \cite{rumigan} & Generator for positives. Handles imbalance & Only Binary classes. Upto 5-10\% imbalance. Tested on images.\\
\hline
GenPU \cite{genpu}& PU classifier and positive generator. & Only Binary classes. Tested on images. Needs positive generator. \\
\hline
CatGAN \cite{catgan}& Multi-class discriminative classifier&  Class densities should be equal. Tested on images. \\
\hline
LSTM-AAE \cite{nips_lstm,anomaly-aae,kdd-autoencoders,vae_deepanshi}& Binary classification. Handles longitudinal data & PoC attempted. Results are not promising.\\
\hline
\multirow{2}{*}{AnoGAN/ALAD \cite{anogan,icdm}}& \multirow{2}{*}{Anomaly detection GAN. Unsupervised.} & Works for upto 10-20\% imbalance.  Assumes  \textit{Binary} classes. \\
&&Does not leverage positive samples. Low recall on anomalies. \\
\hline
\end{tabular}
  \caption{Intersection of Semi-Supervised GANs, PU learning and Anomaly Detection}
  \label{tab:related}
  \end{table*}

\textbf{PU and One Class classification:} Most widely cited initial PU learning approaches \cite{liu2002,lilearning} involve isolating a set of so-called Reliable Negatives (RNs) from the unknown data set. \cite{lee2003} offers a better solution by treating U data to be N data with a decayed weight. However, classifiers trained based on above approaches suffer from a systematic estimation bias \cite{duplessis,kiryo2017}. We realized that one class classification approaches \cite{occ1,occ2,occ3} lack from leveraging the small amount of positive samples available as a ground truth.  In situations where the unlabelled data is growing on a daily basis, instead of learning the distribution of the negative samples as reliable negatives we desire a discriminatory classifier which could preserve information necessary to predict the class label (and become invariant to unimportant details). 

\textbf{Generative vs. Discriminative Clustering:} Prior to building ScarceGAN we reviewed various clustering methods. Generative clustering methods such as Gaussian mixture models, k-means, and density estimation algorithms, directly try to model the data distribution p(x) (or its geometric properties); (2) discriminative clustering methods such as maximum margin clustering (MMC) \cite{xu2004} or regularized information maximization (RIM) \cite{NIPS2010}, aim to directly group the unlabeled data into well separated categories without explicitly modeling p(x). Latter can easily overfit to spurious correlations in the data; especially when combined with powerful non-linear classifiers such as neural networks.  

\textbf{Generative Models:} More recently explored neural networks typically involve training  a generative model; feed-forward neural networks \cite{kingma2014,bengio}, or training auto encoder networks\cite{Hinton504,vincent}.  One problem with such reconstruction based methods is that, by reconstruction, they try to learn representations which preserve all information present in the input examples. This may result in over-fitting. Prior to ScarceGAN, we implemented one such prototype, details in the Section \ref{sec:evaluation}. 

\textbf{GAN Tuning Toolkit:} \cite{lsgan} proposes least squares loss function while training GAN's. It penalizes samples that lie in a long way on the correct side of the decision boundary and generates better samples.  \cite{wasserstein,representationlearning,iantutorial} propose novel techniques and metrics to alleviate concerns (mode collapse, non-convergence) related to GAN training.  \cite{fm} proposes techniques like minibatch normalization, one-sided label smoothing, feature mapping to facilitate better quality generation and prevent generator saturation due to mode collapse etc. In some settings, access to clean labelled data is difficult, \cite{noisylabel} proposes a novel family of GANs called label-noise robust GANs (rGANs), which, by incorporating a noise transition model, can learn a clean label conditional generative distribution even when training labels are noisy. Alternatively, \cite{ccgan} argues that unlike ordinary labels, complementary labels are easy to obtain because an annotator only needs to provide a yes/no answer to a randomly chosen candidate class for each instance. We do suffer from the problem of weak/noisy labels for the negative class, however our prior is very small and their densities are non-uniform. Instead, we leverage this insufficient and noisy truth to our advantage via the \textit{leeway} term. 

\textbf{SOTA at the intersection of semi supervised GAN and PU Learning:} (SOTA - State Of The Art) There has been some interesting work in using GAN for generating samples from an under-represented class. CGANs \cite{cgan} and ACGANs \cite{acgan} have been successful in medical image data augmentation applications, where a converged generator is used to output samples from an under-represented class. DCGANs \cite{representationlearning} demonstrated that GANs can learn a distributed representation that disentangles the concept of gender from the concept of wearing glasses. \cite{rumigan} also attempts to model unbalanced class distribution problem and trains a generator produce samples from a positive/rare class by learning what must be avoided, thereby posing a case for complementary learning. All the these approaches are specifically targeted towards building a robust generator. Generative positive-unlabelled learning (GenPU) \cite{genpu} involves training semi-supervised GANs with a mixture of positive, negative, and unlabelled samples, with the goal of obtaining a classifier that separates the positive unlabelled samples from the negative ones. Unlike the standard GAN, where the optimal generator is of ultimate interest, in \cite{ccgan,genpu} the end product is the optimized discriminator. Our ScarceGAN framework is inspired from these works. We also studied ALAD and AnoGAN ~\cite{anogan,icdm} which use GAN variants for anomaly/outlier detection. While both focus on  positive/negative separation, leverage no prior. We find this as a limitation as the only available ground truth is being overlooked. This lacking is evident in the weaker recall numbers on the positive class, as will be discussed in Section \ref{sec:evaluation}. Table \ref{tab:related} summarizes the options we had and what necessitated building of ScarceGAN.

\section{ScarceGAN Formulation}
\label{sec:scarcegan}
\subsection{Background: GAN and SSGAN}
Generative adversarial networks, originally proposed by Goodfellow et al. \cite{gan}, are unsupervised deep learning machines. It consists of the Discriminator and the Generator networks working in tandem. The two networks are trained through an objective function that implements a two-player zero sum game between a discriminator D – a function aiming to pull apart real from fake input data and a generator G – a function that is optimized to generate input data (from noise) that “fools” the discriminator. 
The discriminator is rewarded for correct classifications and penalized for the other and the generator is rewarded for generating examples that could fool the discriminator. Both models are then updated and the game continues, until the convergence criteria or defined number of learning steps.

\textbf{SSGAN:} Semi supervised GANs or SSGAN for short, is an extension of GAN architecture to address a predictive modeling problem in which there may be a very large dataset of examples (unlabelled), but only a small fraction have target labels (labelled prior). The model must learn from the small set of labeled examples and somehow harness the larger dataset of unlabelled examples in order to generalize to classifying new examples in the future. The discriminator has two path's: a) discriminator in the unsupervised path which leverages the unlabelled dataset as "Real" samples and learns to identify them against the generated "Fake" samples and b) supervised classifier which learns the decision boundaries between the various classes in the labelled prior. Since both the path's share a common base network the weights of the discriminator update during both the training cycles; supervised and unsupervised. One of the early references for the SSGAN architecture can be found here \cite{odena,salimans}

In the supervised discriminator the input flows through the base discriminator model and finally a softmax activation is applied to classify the input into one of the K classes, as provided in the labelled prior. In the unsupervised path the input flows through the same base layers and finally pass through a sigmoid activation layer to classify the data as real or fake.


\subsubsection{\bf{SSGAN Formulation}}
The generator transforms input noise Z $\sim$ $p_{z}$ , typically a standard multivariate Gaussian, to the output G(z) with distribution $p_{g}(x)$. The unlabelled real data is sampled from an underlying distribution $p_{d}(x)$ . The discriminator D(x) predicts the probability of its input coming from $p_{d}$. This is formulated as a min-max game between the generator and the discriminator:
\begin{equation*} \label{min-max}
\underset{p_{g}}{min} ~ \underset{D(x)}{max} ~ \mathbb{E}_{x \sim ~p_{d}} \lbrack ~ \log D(x) ~\rbrack ~ + ~ \mathbb{E}_{x \sim p_{g}} \lbrack \log (1~ - ~D(x)) \rbrack 
\end{equation*}
where the optimal discriminator $D^{*}(x)\equiv \frac{p_{d}}{p_{d} + p_{g}}$ is the one that measures the odds of sample coming from labelled dataset, and the optimal generator was shown to be the minimizer of the Jensen-Shannon divergence between $p_{d}$ and $p_{g}$. 
We define $y_{\mathrm{pred},i} = P_{\mathrm{model}}(x_{i})$, where $x \in \{p_{d}, p_{g}\}$. Here, $P_{\mathrm{model}}$ refers to the final output of discriminator, which is different for supervised and the unsupervised path, due to varied activation functions in the output layer. Let us assume that $P_{\mathrm{model}}(y_i \in C_{y_i})$ is the probability of input sample $x_i$ belonging to its correct class represented as $C_{y_i}$. $y_{\mathrm{true},i}$ is the actual categorical class probability distribution of the sample. 
With this, we define Categorical Cross Entropy (CCE) as:
\\
\[
\text{CCE}(y_{\text{true}}, y_{\text{pred}}) = -\frac{1}{N} \sum_{i=1}^{N} \log P_{\text{model}}[y_i \in C_{y_i}]
\]

The discriminator in the unsupervised path and the generator, both operate on Binary Cross Entropy loss (BCE) with the max-min objective mentioned above. The supervised discriminator uses CCE losses on the known classes provided in the prior.

\subsection{ScarceGAN Formulation}
\textbf{Labelled Data (prior):} We have articulated our unique challenges in Section \ref{sec:intro}. Unlike categorical semi-supervised network proposed in \cite{catgan}, our negative class consists of many sub-classes with quite dis-proportionate marginal densities across these sub-classes. We generate a weak prior by assigning known negative samples into three distinct classes: Dormant players with distribution $p_{d}^{D^{-}}$, Normal players with distribution $p_{d}^{N^{-}}$ and Heavy players distribution $p_{d}^{H^{-}}$. These class categories are referred to as `D', `N' and `H' respectively. Class allocations are done based on various game play and engagement statistics of these players. More details on the choice of classes will be discussed in section \ref{sec:evaluation}

\begin{figure}
  \includegraphics[width=0.45\textwidth, height=5cm,keepaspectratio=true]{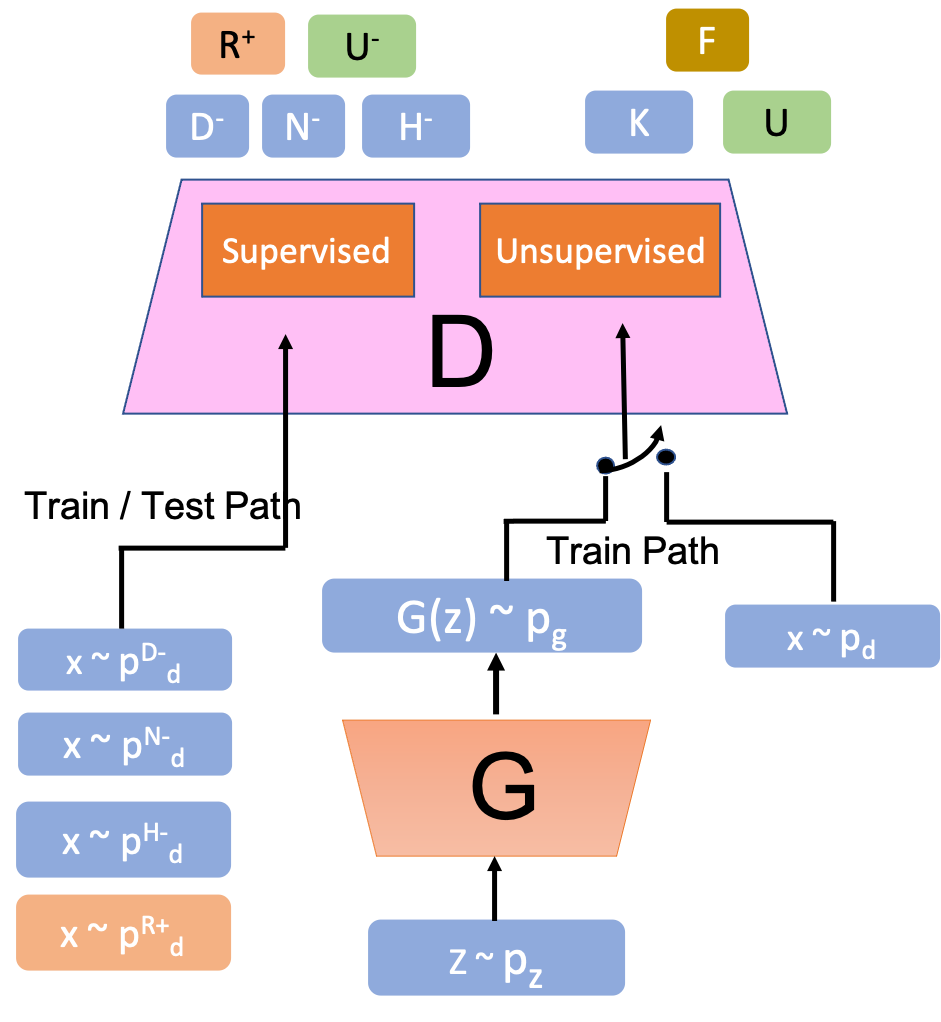}
  \caption{ScarceGAN Architecture}
  \label{fig:scarcegan}
\end{figure}

The negative labels are weak and the sample dataset is only a partial representation of the negative sample space. Unlike the case in \cite{catgan,rumigan} where the prior samples are correctly labelled and in most cases a complete representation of the distribution for their respective classes. Hence, it is quite possible that due to noisy/weak labels and lack of feature space distribution completeness, finding a good decision boundary for negative classes might be a problem. But, negative samples are in abundance and the unsupervised path provides a premise to comprehend its learning. Hence, in ScarceGAN, we define a Leeway class called the Unknown label class , namely `U', with the distribution $p_{d}^{U^{-}}$, which is a part of the negative class. This class accommodates players which are definitely not risky but not confidently fitting into any of the 3 negative classes. The Fifth class is the risky players' class and is denoted as `R', $p_{d}^{R^{+}}$. This is the positive class.

\{$p_{d}^{D^{-}}$, $p_{d}^{N^{-}}$, $p_{d}^{H^{-}}$,  $p_{d}^{R^{+}}$\} are the distributions of the various classes in $p_{d}$, as the supervised input.  In the unsupervised path the unlabelled data is simply represented as $p_{d}$. The $p_{d}^{U^{-}}$ is the unknown distribution which the discriminator explores and provides as a complementary output. With this we are now ready to formalize ScarceGAN framework. Figure \ref{fig:scarcegan} depicts ScarceGAN architecture.

\subsection{ScarceGAN Discriminator}

\subsubsection{\textbf{Supervised Path:}}
This path is trained with labelled data. The input data labels are of four types. To the output layer of this path a softmax activation is applied with $5$ outputs. 4 of these correspond to the 3 known negative and 1 positive class and the $5^{th}$ class corresponds to the `unknown' negative class. We now discuss how the objective functions are structured for this arrangement. 
\\
\noindent\textbullet\ \textbf{Objective 1:} \textit{High Confidence Prediction and High Recall for Positive class samples:}
This is achieved via two constrains: 1) Penalty: Categorical Cross Entropy (CCE) between predicted and true, for labelled samples coming from positive class and 2) Reward: percentage recall on positive samples. We can then write the combined loss of the supervised discriminator on the positive samples as:
\begin{equation*} \label{sd_R}
L_{D}^{S^{+}} = CCE( x | C_y \in R) + \%Recall^{+ve} 
\end{equation*}
\vspace{-0.5cm}
\begin{equation*} 
L_{D}^{S^{+}}= - \frac{1}{|p_{d}^{R^{+}}|} \sum_{i=1}^{|p_{d}^{R^{+}}|} \log P\it{model}[y_i \in R ] + \%Recall^{+ve}
\end{equation*}
\\
 Here instead of looking at the overall CCE, as in the traditional SSGAN formulation with balanced labels, we split it into two parts. As a part of objective 1 we are looking at specifically reducing the CCE for true positive samples. Which is synonymous to penalizing low confidence predictions. The second term \%$Recall^{+ve}$, is the \% of the correctly classified positive samples out of the total provided in the batch. The precision on positive class is handled indirectly by objective 2, which is on the negative class samples.
\\
\noindent\textbullet\ \textbf{Objective 2: }\textit{High Confidence Prediction for known Negatives and `leeway' for unknown negatives :} Leeway criterion, to our knowledge is first of its kind. As earlier stated, there are many patterns that we missed (small prior) and the existing labels themselves might be weakly assigned (noisy labels). Through the leeway term we allow the classifier to say that it isn't confident about their exact negative class but for sure they are not positives. Because, we are sure that these samples from the prior indeed belong to the negative class. This is achieved by the following cost objective.
\begin{equation*} \label{sd_N}
L_{D}^{S^{-}} = \alpha ~CCE( x | C_y \in \{D,~N,~H\}) + ( 1 - \alpha)~CCE( x | C_y \in U)
\end{equation*}

Here $\alpha$ and ( 1- $\alpha$) are the dynamic weight parameters attached to the losses associated with the known negative and unknown negative subsets, respectively. Such that $\alpha$ + ( 1-$\alpha$ ) = 1.0. Since the $y\it{true}$ for the negative class only indicates their respective known negative class labels, the logits of all labelled inputs are always zero for the unknown class output position in the supervised discriminator path. While computing this custom loss function, we force assign a label `U' to all the negative samples, in the (1- $\alpha$) path. Hence CCE is calculated twice for all the negative class samples and weighted average is taken. 

When $\alpha$ = 1, the loss latches on to high confidence on exact classification of all the negative samples.  We find the setting between $0.7$ - $0.6$ ideal for $\alpha$ in this setting, where we believe about 30\%-40\% of the negative labels might be noisy. Combining the two, the loss of the discriminator in the supervised path is defined as: 
\begin{equation*} \label{sd_N}
L_{D}^{S} = L_{D}^{S^{+}} + L_{D}^{S^{-}}
\end{equation*}

\subsubsection{\textbf{Unsupervised Path:}}
This path in-directly trains the discriminative classifier with unlabelled data ($p_{d}$). The leeway term introduced in the supervised path is important here as well. The input data has no labels. In traditional SSGAN architecture, the unsupervised path terminates with a `sigmoid' activation function. 

In ScarceGAN formulation, the output layer of this path a softmax activation  with 3 possible outcomes, namely: 'K', 'U' and 'F'. K corresponds to Known labels (can be any from D, N, H or R), U corresponds to Unknown (from leeway term) and F is for the samples identified as Fake (ideally the generated ones). Hence we do not use the BCE loss but instead do the following change in the cost objectives:

\begin{equation*} 
L_{D}^{US^{Real}} = \alpha~CCE( x | C_y \in {K}) + ( 1 - \alpha)~CCE( x | C_y \in {U}) 
\end{equation*}
\begin{equation*} 
L_{D}^{US^{Fake}} = CCE( x | C_y \in F)  
\end{equation*}
\begin{equation*}
L_{D}^{US} = 0.5 * (L_{D}^{US^{Real}} + L_{D}^{US^{Fake}})
\end{equation*}

In the first statement, $L_{D}^{US^{Real}}$ we calculate the loss due to unlabelled samples (real) which are to be split between the \textit{K} class or the \textit{U} class. $\alpha$ and 1-$\alpha$ are dynamic and complementary weights, with $\alpha$ + ( 1-$\alpha$ ) = 1.0. Since we do not know the actual labels, we calculate their cross entropy on being from known `K' or unknown `U' class. Hence the true class $C_y$ is first set to $K$ and thereafter to $U$. The gradients backpropagate to adjust/update weights of the intermediate shared layers to minimize the penalty by best exploiting the leeway label, U. In other words, the leeway term offers a loss exemption proportionate to the value of the term (1 - $\alpha$).

In the $L_{D}^{US^{Fake}}$, the expected label is fixed, `F'. so CCE is directly computed on this label. As with the traditional GAN's, we average out the total loss after training the unsupervised discriminator on real and fake data one after another, starting with the real data first.


\subsection{ScarceGAN Generator}
\cite{badgan} establishes and systematically proves that semi-supervised learning methods based on GANs, cannot obtain good classification performance and a good generator at the same time.  It theoretically shows that given the discriminator objective, good semi-supervised learning indeed requires a bad generator, and proposes generator objectives to achieve that. This aspect of generator learning is called ``complementary learning'', where it learns in the complementary space, in our case specifically complementary to all 5 classes ( D, N, H, R and U). 

\noindent\textbf{Why Bad Generator in ScarceGAN?} Techniques proposed in \cite{badgan,fm} target to exploit the underlying manifold structure of the real data samples. Here the objective of the generator is to match the first-order feature statistics between the generator distribution and the true distribution. After understanding the underlying manifold the bad generator produces the complementary samples. The optimized discriminator obtains correct decision boundaries in high-density areas in the feature space if the generator is a complement generator. We implemented the following loss function in the generator as prescribed in \cite{badgan} and indeed observed a consistent 15\%,10\% boost in precision,recall of ScarceGAN respectively:
\begin{equation*}
\begin{split}
\underset{min}{G} ~ -H(p_G) + \mathbb{E}_{x \sim p_{g}} & ~\log (p(x)) \nparallel \lbrack p(x) > \epsilon \rbrack + \\
 & \|\| \mathbb{E}_{x \sim p_{g}} f(x) - \mathbb{E}_{x \sim p_{d}} f(x) \|\|^2
\end{split}
\end{equation*}
The first term tries to increase the entropy of the distribution of generated features. This improves the diversity of the generated samples in the feature space. As suggested in the paper we implement this by a pull-away term which tries to orthogonalize the features in each  generated batch by minimizing the squared cosine similarity. The second term, indicates enforcement of sample generation in the low density input space. This is done by minimizing the expectation of the discriminator's confidence on the class of the generated sample, where $\nparallel \lbrack \cdot \rbrack$ is the indicator function. The confidence threshold ($\epsilon$) is chosen to be $0.75$ in the current deployment. The third term, is the feature matching term proposed in \cite{fm}. Here, f() denotes activations on an intermediate layer of the discriminator. We let the generator spy on the intermediate layers of the discriminator to understand features that are most discriminative of real data. We train the generator to match the expected value of the features on an intermediate layer of the discriminator. All put together gives a bad generator which boosts the capability of the discriminator to learn well on the unlabelled samples. 



 

\section{Handling longitudinal data}
\label{sec:method}

 Data to be analyzed is multi-dimensional, consisting primarily of player's game play,  platform transaction and click stream information. This data is available at each interaction of the player with the platform. From the multiple available dimensions we select features along three meta dimensions of immoderation - Time, Money and Desperation. Table \ref{table_features} highlights some of the features. 
 
 Most of the anomaly detection and rare class identification work also has access to such a longitudinal data stream. However, almost all of the publicly available datasets like \cite{kddcup,kaggle} have been aggregated and published. We have found that trend preserving aggregations, like the one we discuss in this section are much more effective than longitudinal aggregates like the mean, max and median values of the time series data.

\begin{table}[t]
  \centering
  \footnotesize
  \begin{tabular}{ | c | l |c| }
\hline
Dimension & Feature & Frequency\\
\hline
\multirow {4}{*}{Time} & Count of games played & daily \\
& Count of weekend games & weekly\\
& Count of weekday games & weekly\\
& Count of late night games & daily\\
\hline
\multirow {4}{*}{Money} & Loss & daily \\
& Count of Add Cash transactions & daily\\
& Count of add cash failures & daily\\
& Number of distinct payment modes & daily\\
\hline
\multirow {3}{*}{Desperation} & Win Ratio & daily \\
& Count of Invalid Declarations & daily\\
& Total bad quality hands played & daily\\
\hline
\end{tabular}
  \caption{Sample Game Play Features}
  \vspace{-0.3in}
  \label{table_features}
\end{table}


\subsection{Feature Synthesis:}
\cite{shapelets_anomaly} looks at Dynamic Time Warping with shapelets on time series data to synthesize derived features to detect addicted players. Many works have leveraged LSTM and BiLSTM networks for time series aggregations for their classification \cite{lstm2,nips_lstm}. Our initial analysis and experimentation with DTW and shapelet as well as LSTM based features did not yield promising results (more details in Section \ref{sec:evaluation}). 
\\
\noindent\textbf{Third Order Statistics over Longitudinal Data:}
Longitudinal data contains history of how the feature has been changing over time. One of the ways to succinctly represent a time varying pattern is to model the underlying time series and use the hyper-parameters of the model as features.  Since these are the hyper-parameters of a model, we call them as third order statistics as opposed to those directly composed over the series like the shapelets, WEASEL and DTW \cite{shapelet,weasel,dtw}.  We finalized on open-source library for uni-variate time series forecasting called \textit{Prophet} \cite{prophet} from \textit{Facebook}$^{\textregistered}$\cite{Facebook} after experimenting with ARIMA \cite{arima} and TBATS \cite{tbats}.  

\begin{figure*}
 \centering
  	\begin{subfigure}[b]{0.33\textwidth}
         \centering
         \includegraphics[width=\textwidth]{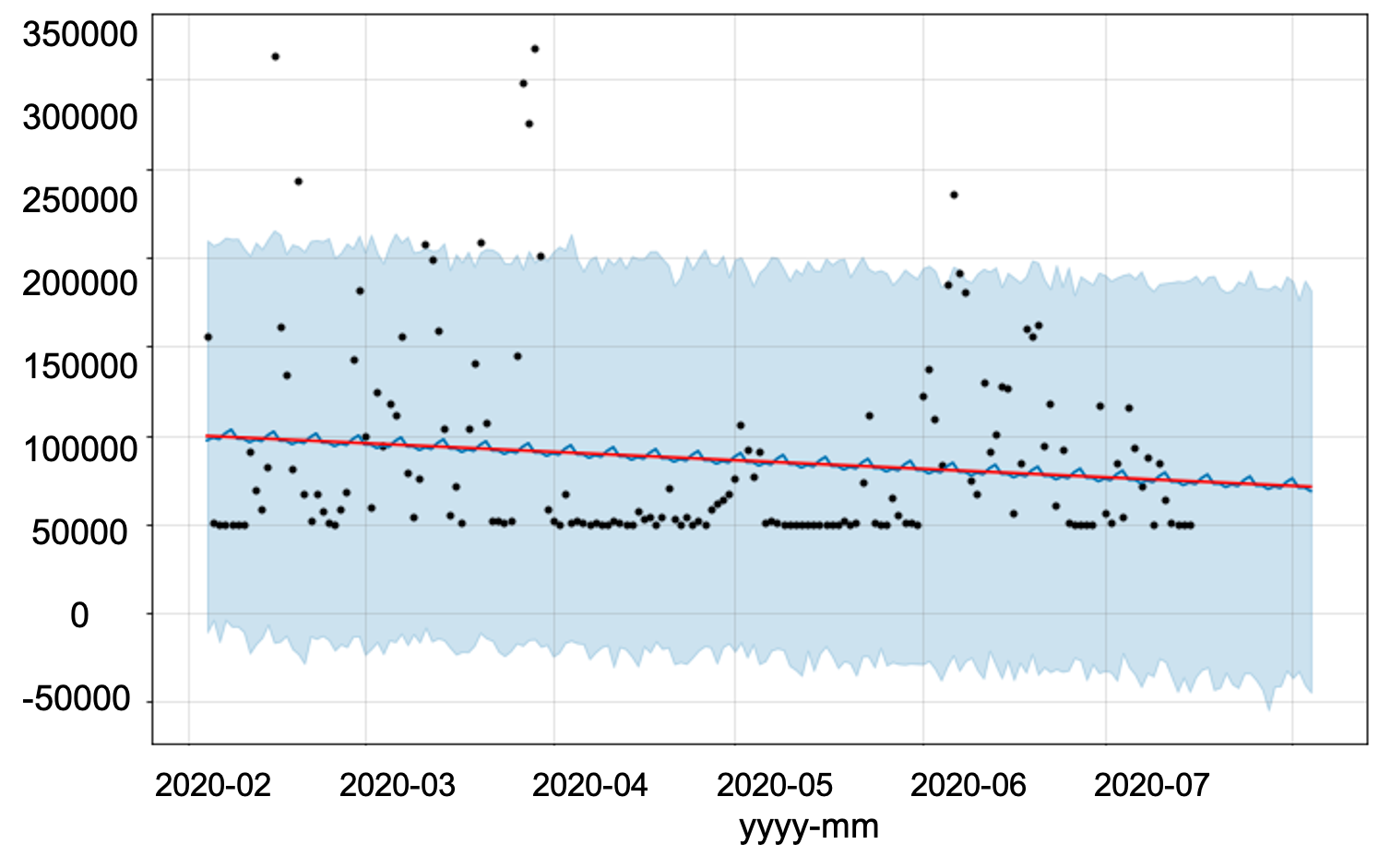}
         \caption{with trend (A) + seasonal (B)- original}
         \label{sub:original}
     \end{subfigure}
     \hfill
	\begin{subfigure}[b]{0.33\textwidth}
         \centering
         \includegraphics[width=\textwidth]{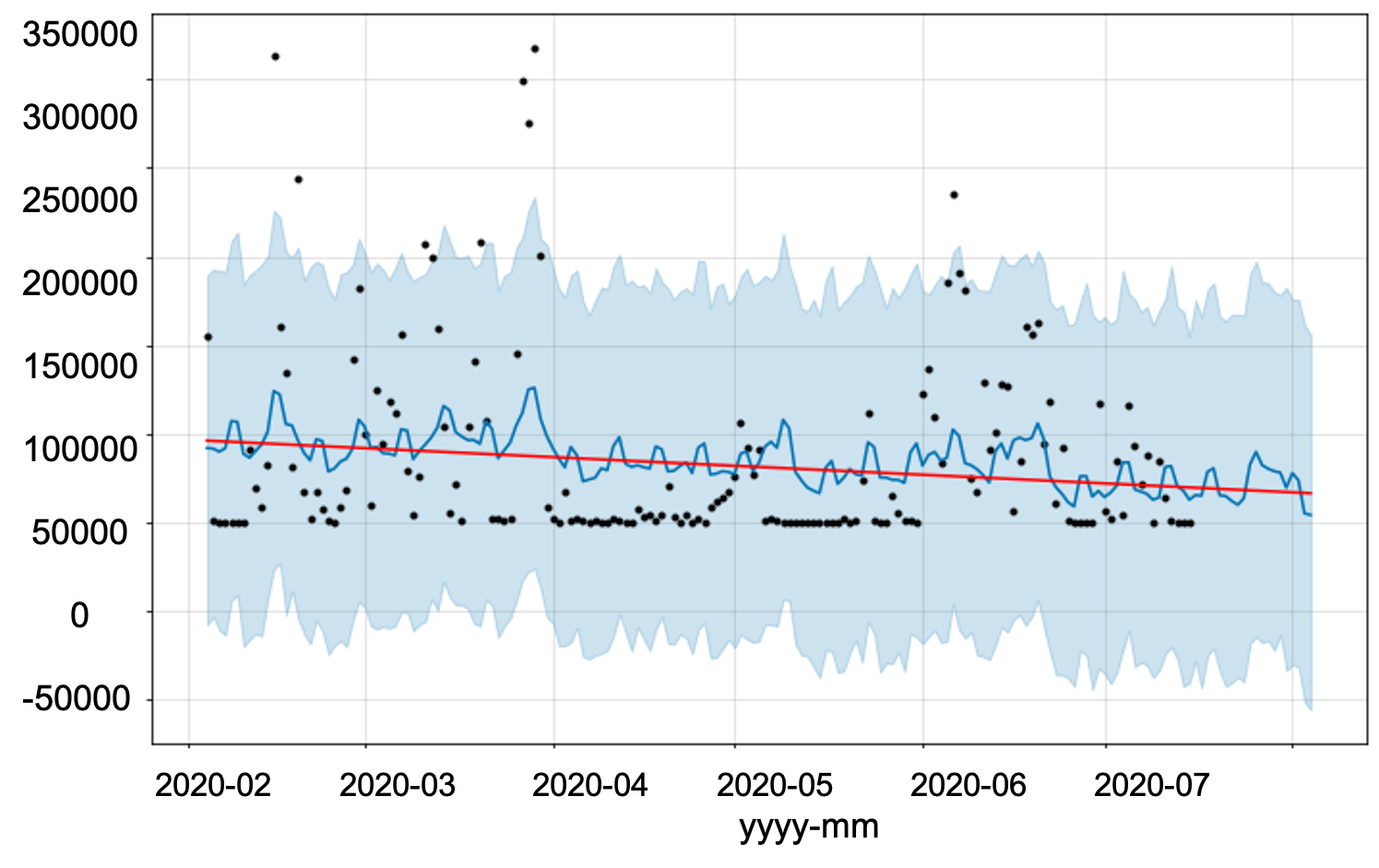}
	     \caption{with A + customized seasonal(C)}
         \label{sub:seasonal}
     \end{subfigure} 
     \hfill
  \begin{subfigure}[b]{0.33\textwidth}
         \centering
         \includegraphics[width=\textwidth]{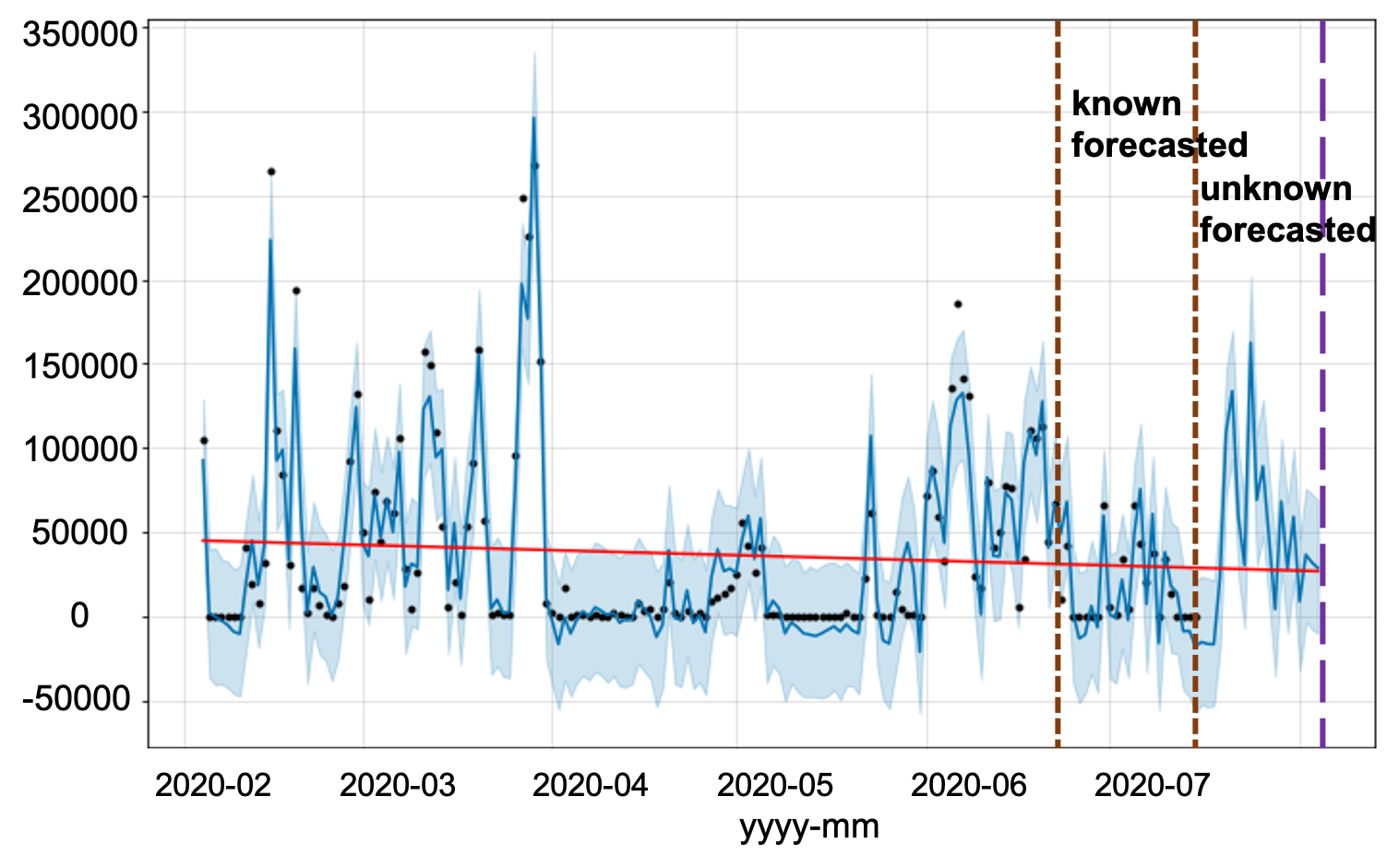}
	    \caption{with A + C + custom regressor}
         \label{sub:regressors}
     \end{subfigure} 
     \caption{Feature Time Series Modelling using Customized Prophet: dots - actual samples, blue line - predicted values, red line - trend line}
       \vspace{-0.1in} \label{sub:prophet}
\end{figure*}

We believe that each player's pattern has a predefined notion of a `normal' engagement on the platform. This is quite synonymous to pattern of read and write operations for a given application workload. Prophet establishes a notion of piece-wise linear trend and allows the user to specify expected change points in the series. 
We found that the first three power spectrums of the seasonal component's periodogram captured the longitudinal variation much better compared to the vanilla prophet tuning (Figure \ref{sub:original} vs. \ref{sub:seasonal}). Also, as shown in Figure \ref{sub:regressors} using the concept of a `regressor' which denotes an independent parameter associated with the time series, for instance, the day of the week, further boosts model's prediction on the future values. 
We observe that custom tuning of prophet's parameter grid helps in quite efficient forecasting. We observed Mean Percentage Errors of 5-10\% on forecasted predictions.

We define our own set of $10$ hyper-parameters from Prophet's parameter grid, which includes $1^{st}$, $2^{nd}$ and $3^{rd}$ harmonics of the seasonality periodogram, Laplace smoothing coefficient, mean and standard deviation of the series' change rate and the growth rate. 
These third order statistics, were calculated over each of the feature time series (Table \ref{table_features}). The length of observation per sample varied from $15$ upto $45$ days. These third order statistics per feature formed an input to ScarceGAN. It is not mandatory to feed model hyper-parameters to ScarceGAN. As we will see in Section \ref{sec:evaluation}, ScarceGAN can operate with longitudinal aggregations as features too.


\vspace{-0.1in}

\section{Empirical Evaluation}
\label{sec:evaluation}
In this section we present empirical evaluation of the ScarceGAN to identify rare class samples from  multiple sets containing large volumes of unlabelled samples. We compare contributions of ScarceGAN over various variants of the approach that we considered and a variant from of the state of the art, auto-encoder based prototype. Thereby, we establish the robustness and accuracy of ScarceGAN. We later extend ScarceGAN network as-is to evaluate its potential on detection of intrusions in the KDDCUP99 \cite{kddcup} data for specifically rare classes, F2L and U2R, with an imbalance of 0.093\% and 0.005\% respectively. We leverage semi-supervised approach of ScarceGAN as opposed to the existing supervised learning approaches for this dataset and report a new record of F1-score for one these classes while remaining competitive in the other. Further, scarceGAN outperforms the recall benchmarks established by recent GAN based unsupervised models for the positive imbalanced class identification for the same dataset.
\vspace{-0.1in}
\subsection{Datasets:}
We built our labelled prior using players' data observed at different timelines in the platform. The three distinct observation intervals were the first, the third and then the last quarter of the year 2020. We collected over a million samples for these intervals in total. Our input consists of the longitudinal feature patterns for at least 15 and at most 45 active play days on the platform. Using data from the business database, our prior on negative class dataset samples was finalized. We created about 500 samples per negative class out of the million samples. Rest of the samples were retained as unlabelled real data. Data span quite proportionately across the three chosen quarters. 

Ground Truth Subset Creation: Predicted Risky users were 1) either handpicked by the business via various monitoring and reporting statistics or 2) by the rule engine which uses parameters similar to one mentioned in the Table \ref{table_features}. The rules are derived based on percentile level thresholds on the parameters to identify anomalies followed by a combination logic from domain understanding. All the players are assigned scores on these dimensions and those surpassing the thresholds on certain number of dimensions are flagged. These predicted users are then sent for counselling and they are either 1) predicted as false positives and are allowed to continue playing as usual, 2) predicted emerging risky and some kind of throttles are placed on their engagement on the platform or 3) predicted highly risky and are barred from playing. There is no way to measure recall, as the complete ground truth is unknown, however the precision of the existing flagging mechanism fluctuates between 7\% - 15\% on a monthly basis for the past 6 - 8 months. Our verified ground truth is obtained from confirmed risky players which are either controlled or abstained from playing. 
\vspace{-0.1in}
\subsection{Model Training and Test Set Description:} 

\textbf{Operations and Activations:} ScarceGAN's source code is available at \cite{github}. Very briefly, our discriminator's base networks are fully connected layers with leaky relu activations. The unsupervised network applies softmax activation with 3 outputs and the supervised with 5 outputs. Generator has a single fully connected layer with leaky relu, followed by batch normalization layer feeding to a final fully connected layer with relu activation. Our mini-batches are configured to 32 with equal samples over 4 classes for the supervised path. We use LR schedule with decayed learning rate and adam optimizer with $\beta1$ of $0.5$ and $\beta2$ of $0.9$. We leverage some relevant tips and tricks mentioned in \cite{iantutorial} for layer initialization and regularization.
\\
\noindent\textbf{Labelled Test Data:} We had $500$ each labelled samples from the negative classes (D ,N, H). Train set was re-sampled from $350$ fixed samples on every run, while the held out (test set) was fixed to remaining $150$ odd samples. These test samples  were never introduced in any of the training runs. Out of these $350$, $200$ samples each were randomly selected for training supervised ScarceGAN. We had $202$ samples for the positive class. We selected $150$ fixed samples for training and remaining $52$ were fixed for testing likewise.

\noindent\textbf{Unblocked Players:} As aforementioned, out of the total players sent for counselling for the past 6-8 months about 85\% or higher were declared as false positives post counselling. We had access to about $186$ such samples. We evaluate ScarceGAN and our other prototypes on these samples to assess the verbosity of the model against FP's.

\noindent\textbf{Self Identified Users (SIU's):} The $202$ positives were flagged by rule engine or directly by the business. Additionally we had access to a few positive samples where the players self identified themselves as over-indulged and requested for counselling and were verified risky. We did not expose these samples to ScarceGAN and other prototypes in the training phase and directly tested their recall on the trained model.

\noindent\textbf{Set 1, Set 2, Set 3:} We evaluate our model with held out test and unblocked user patterns. This data repository is built from samples collected over different time periods. Set 1, 2 and 3 represents samples eligible for positive prediction on three distinct days on the online skill gaming platform. Set 1, 2 and 3 belongs to March $1^{st}$ of 2021 , November $12^{th}$ and   August $14^{th}$ of 2020 respectively. One of the reasons behind choosing these dates being that on each of these days we had a significant number of positive flags which were confirmed from counselling. \% Recall indicates, recall on these confirmed positives, \% verbosity indicates the total volume of flagged positive samples by the model. Beyond what was verified, we cannot comment on the precision/recall of the other samples as they were not flagged by existing rule engine and hence some of these could be actually positive samples. 

\noindent\textbf{Precision Recall Interpretation:}  ScarceGAN and its variants are tasked with identification of the rear/scarce positive samples. Hence, our precision recall is purely on positive class verses the entire negative class. In the held out test dataset we have about a total of $450$ samples from the negative ($150$ x 3) and $52$ samples from the positive class.
\vspace{-0.1in}
\subsection{Results:}
\subsubsection{\textbf{Comparison with Anomaly Detection SOTA- Auto Encoders:}} Challenges associated with the identification of positive samples draw a parallel with the domain of unsupervised anomaly detection. Some recent competitive works \cite{nips_lstm,anomaly-aae,kdd-autoencoders,vae_deepanshi} suggest an Adversarial Auto Encoder (AAE) based framework to model `normal' class distribution and use high reconstruction losses on the trained decoder to mark as anomaly. Prior to building ScarceGAN, we prototyped an AAE based solution and used Prophet features as discussed in Section \ref{sec:method} as input. AAE is a combination of GANs and Auto encoders (encoder and decoder network). The two networks when trained in tandem, learn the complex distribution of the negative samples in the latent space. The training data is the same unlabelled samples used for training ScarceGAN.  

\begin{figure}[h]
\includegraphics[width=0.45\textwidth]{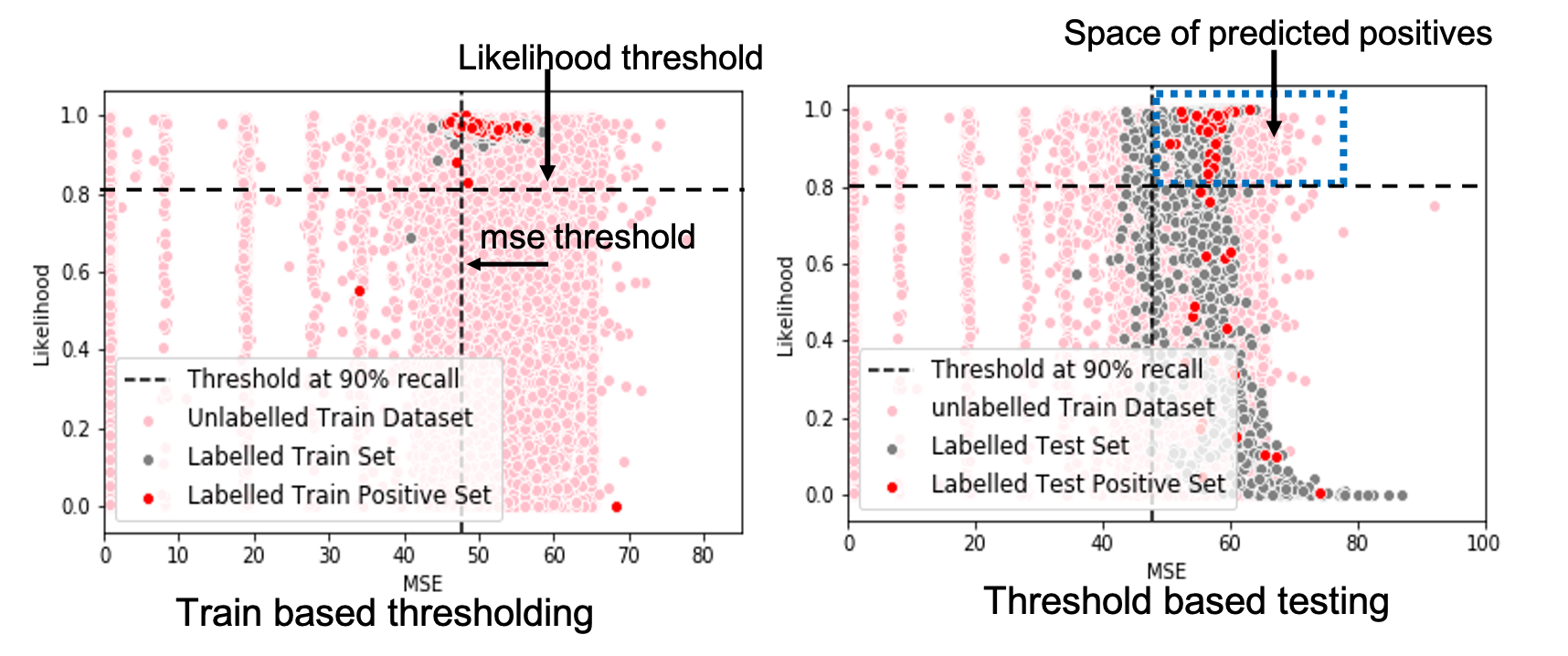}
\caption{AAE based positive detection: MSE and Likelihood thresholds cannot escape verbosity. Highly overlapping distribution between positives and negatives in our case.}
\label{fig:mse_likelihood_plot}
\vspace{-0.1in}
\end{figure}

The auto encoder part of the network is trained to minimise the reconstruction loss (KL divergence) between the encoder and the decoder. We denote this error as `mse'. In the regularisation phase, the adversarial network updates the discriminator to distinguish between the true samples (from generic standard normal distribution) and the generated encoded samples (from the encoder). This model captures potential positives using two different controls - the \textit{likelihood} of an anomaly (discriminator identifies the sample as Fake) and the reconstruction error. We used the labeled samples to set this likelihood/reconstruction error threshold on the trained model as shown in Figure \ref{fig:mse_likelihood_plot}.


Two types evaluations done using this AAE model. The unsupervised train network is the same in both. However, the thresholds (mse/likehood) are differently set. The first model (AAE - 90\% recall) is finalized as follows: thresholds are assigned such that at least 90\% of the positive samples in the train labelled set are recalled. As expected all the positive samples in the test set are recalled but with a only 5\% precision. The second model, we try to reduce the verbosity by fixing the precision. Unfortunately, we observe that samples from the four training classes are so intermixed that the best obtainable precision is only 50\%. This precision of 50\% on trained labelled samples drops to 6\% in the test set with a recall of 55\%. Results for Set 1, 2, 3 and unblocked, SIU's are presented in Figure \ref{detailed_results}. We see that as we focus on recall, the verbosity goes for a toss. Though Recall is our primary objective, high verbosity will warrant a another second level classifier. Figure \ref{fig:mse_likelihood_plot} highlights how the thresholds fixed on the basis of the labelled train samples to retain over 90\% recall in positive (risky) increases the overall verbosity of predicted positives (dotted box) in train and test sets.

We conclude that a completely unsupervised approach based on AAE does not meet our dataset classification requirements. As discussed in literature, AAE based techniques try to understand the structure of the class with focus on minimization of KL divergence. However a more discriminative approach might be suitable. 

\subsubsection{\textbf{Performance of ScarceGAN and Variants:}}

\begin{table}[t]
  \centering
  \footnotesize
 \begin{tabular}{||c| c| c||}
 \hline
  Model &  Precision(mean)\%  & Recall(mean)\% \\
 \hline  
  ScarceGAN & 75.7 & 72.5 \\
  Vanilla SSGAN & 34 & 24 \\
  ScarceGAN 2 class & 37 & 60  \\
  ScarceGAN w/o Bad Generator & 60 & 62 \\
  ScarceGAN w/o Leeway & 40.5 & 57 \\
  (SOTA) AAE - Recall focused & 5.3 & 100 \\
  (SOTA) AAE - Precision focused & 6.4 & 55.7\\
  \hline
\end{tabular}
 \caption{Performance of ScarceGAN and Variants on Held Out Set} 
 \vspace{-0.2in}
 \label{verify_variants}
\end{table}

\begin{table*}[t]
  \centering
  \footnotesize
 \begin{tabular}{|c|c|c|c|c|c|c|c|c|}
 \hline
 \multirow{2}*{Model} & Unblocked (186) &         \multicolumn{2}{c|}{Set 1 (78,388)}         & \multicolumn{2}{c|}{Set 2 (112,692)}            & \multicolumn{2}{c|}{Set 3 (113855)} & SIU's (7)\\\cline{3-8}
                   & False Positives                 & Verbosity & Recall         & Verbosity & Recall          & Verbosity & Recall         & Recall \\
 \hline  
  ScarceGAN (avg)       & 107 (57.5\%)           & 1242(1.5\%) & 3/4(75\%)       & 6902 (6\%)           & 8/9(89\%)          &3397(2.9\%)   & 10/11(91\%)        &6/7(86\%) \\ 
  Vanilla SSGAN (avg)   & 67  (36\%)            & 0(0\%)      & 0/4(0\%)        & 508  (0.2\%)         & 2/9(22\%)          &2487(2.1\%)   & 2/11(18\%)         &0/7(0\%) \\
  AAE - Recall 90\%(avg)     & 185 (99.4\%)          & 30,477(38.8\%) & 4/4 (100\%)   & 80907 (72\%)         & 9/9(100\%)         &39407(34.6\%) & 11/11(100\%)       &1/7(14.2\%) \\
  AAE - Precision 50\%(avg)  & 129 (69.3\%)         & 1018(1.2\%)   & 1/4(25\%)      & 22731 (20\%)         & 3/9(33\%)          &1446(1.2\%)   & 6/11(54.5)         &0/7(0\%) \\
  \hline
\end{tabular}
 \caption{Performance of ScarceGAN and Select Variants on Unblocked Players, Set 1, 2 and 3 and SIU's} 
 \label{detailed_results}
 \vspace{-0.2in}
\end{table*}

Tables \ref{detailed_results} and \ref{verify_variants} (mean taken over 6 runs) lists the complete results.
\\
\textbf{Vanilla SSGAN:} We first look at the traditional SSGAN with CCE loss on supervised and BCE loss on unsupervised, with all 3 negative, 1 positive samples as real. Generator is evaluated with BCE loss on Fake/Real samples. Here, since positive samples are rare in the unlabelled set, the supervised classifier  essentially unlearns the positive distribution. The precision-recall of the held out set is 34\% and 24\% respectively confirms this lack of knowledge on the positive class. The verbosity of this network on the unblocked users is the least - 36\% . Low verbosity and low recall in set 1, 2 and 3 indicates that, perhaps all of the positive distribution is also classified as negative due to lack of exposure and high imbalance in the unlabelled samples. This necessitates a different formulation over the vanilla SSGAN architecture.

\textbf{ScarceGAN:} ScarceGAN is our proposed solution to solve scarce class identification problem. It demonstrates the highest precision, recall of 75.7\% and 72.5\% respectively on the held out dataset. As similar with other variants, its verbosity on the unblocked players is reduced to 48\%, thereby showing a lift in the precision from 14\% to 48\%. It achieves recall between 85\% to 90\% on already flagged and confirmed positives from a minimum of 80k test samples. The verbosity in positive predictions in Set 1, 2, 3 is the least amongst its class of variants, hovering between $1$\% to $6$\%  of the total assessed. We observe that on the SIU's, the players which self identified themselves, ScarceGAN shows a recall of 85\%, thereby possibly extending the value proposition of the framework.

\textbf{ScarceGAN with binary predictions:} We did evaluate the case where we use only one negative class and a positive class along with the leeway term for the spill over of the less confident negative samples. We validated the need for our design choice of ``Divided Focus''. With binary classification the precision and recall drops to 37\% and 60\% respectively. 

\begin{figure}
  \includegraphics[width=6cm, height=5cm,keepaspectratio=true]{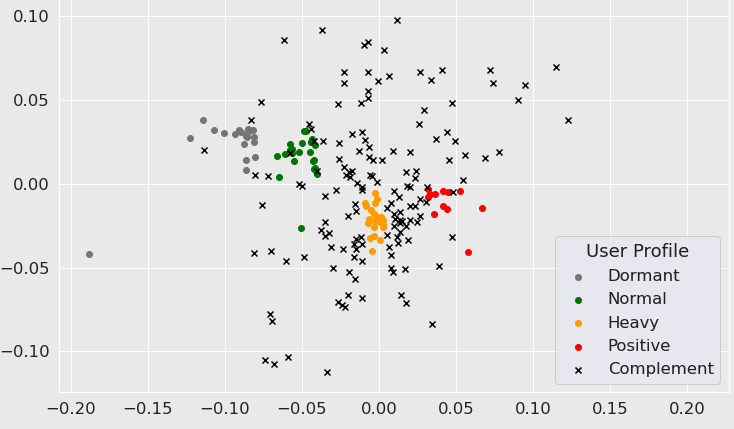}
  \caption{Complementary samples generated by generator}
  \vspace{-0.3in}
  \label{complementary}
\end{figure}

\textbf{ScarceGAN w/o Complementary Generator:} We disabled the complementary generator keeping other portions of the ScarceGAN network intact. Essentially, the generator now used CCE on the fake samples as its loss guidance.  Recall/Precision on test data indicates a significant lift from the Vanilla SSGAN but deterioration by 10\% in the performance against ScarceGAN. Complementary generator specifically helps the discriminator focus on improving its decision boundaries on negative samples. Figure \ref{complementary} shows how the generated samples are indeed complementary. The Generator, very distinctly keeps the generated samples close yet separated from each of the class boundaries. This poses a real challenge to the discriminator in robust understanding of the class decision boundaries. Longitudinal modeling also enhances the separation between positive and negative samples (Figure \ref{kde-pn}(A)).

\begin{figure}
  \includegraphics[width=0.45\textwidth, height=5cm,keepaspectratio=true]{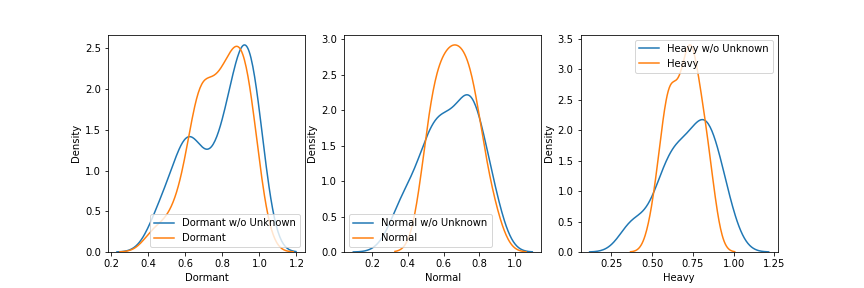}
  \caption{Lift in the confidence of predictions due to Unknown class (leeway)}
  \label{unknown_prob}
  \vspace{-0.1in}
\end{figure}

\textbf{ScarceGAN w/o U Class} Here, we retain all the components of the ScarceGAN except the leeway term which creates a space for the unknown class. Essentially, we are penalizing the discriminator for not respecting the weak negative labels. We expect to see discriminator getting confused and overall reduction in its performance. We observe that model's precision takes a severe hit (40.5\%) and only slightly better to vanilla SSGAN. On the other hand, adding Unknown class as a leeway lifts the precision to 75.7\%. We expect the confidence of prediction of the known negative class naturally lift as an effect of this term. As the low confidence sample will spill over to the U class. Figure \ref{unknown_prob} highlights this effect.
 
\subsection{\textbf{ScarceGAN on Intrusion Dataset:}}
The KDDCUP99 10\% dataset is a network intrusion dataset. The train data has 1,074,992  data samples and the test dataset has 311,029 data samples. The dataset is highly skewed with 4 intrusion and one normal category. About 98.61\% of the data belongs to either the Normal or DoS categories, probe is 1.289\% , R2L is  0.093\% with only 999 samples and U2R  is 0.005\% with only 52 samples out of over a million samples in the train. R2L, U2R, DoS and Probe are intrusion categories. More details here \cite{kdd-data,kddcup}.  

There has been two parallel and fundamental research efforts on this dataset. 1) using Probe, R2L and U2R as specific instances of extreme class imbalance and build anomaly detection classifier to detect these samples and 2) since `normal' class is a minority with only 97K out when compared to total samples of all the intrusion classes, treat normal class as the anomaly and construct precise classifier on detecting the normal class from rest of the intrusion class (treated as one class). 
Both these objectives can be theoretically achieved by ScarceGAN framework as it has been specifically positioned as rare class identification with a small prior. 

\subsubsection{\textbf{Rare Class Identification: Probe, R2L and U2R }} Table \ref{fig-kdd-rare} shows the comparison of the results obtained by ScarceGAN on three rare classes of KDDCUP99. We see that ScarceGAN outperforms the existing F1-score track record for R2L (with 0.093\% imbalance) and produces comparable performance on the other two. Detailed analysis on the only state of the art so far is available here \cite{intrusion2}. 
ScarceGAN configuration for R2L identification: For this task, our positive rare class is R2L and negative class is the normal class as in \cite{intrusion2}. Out of the total 999 samples of the positive class, we use 900 samples as labeled positive prior, and 30k as the negative labelled prior (normal class). Instead of 3 negative prior, we only have one here. The network cost functions remain intact. We only lift the batch size of the training to 64, due to increase in the  total number of labelled samples. The rest of the 67k samples from negative class and 99 samples of the positive class are fed as unlabelled real samples.  Interestingly, though our ScarceGAN accommodates a concept of leeway class, in this setting since prior is strong, the leeway class does not accumulate any samples in the training as well as the test dataset. ScarceGAN reported results are an average over 5 different runs.
Till now specialized anomaly frameworks like AnoGAN \cite{anogan} and ALAD \cite{icdm} have only looked at imbalance between 10-20\%. We see no published results so far on these scarce R2L, U2R and Probe categories.

\begin{table}[t]
  \centering
  \footnotesize
 \begin{tabular}{||c c c c c||}
 \hline
 \multirow{2}*{Model} & \multicolumn{3}{c}{F1-Score}  & \multirow{2}*{Approach} \\
                   & Probe & R2L & U2R &\\
 \hline  
  Naive Bayes & 75 & 18 & 1 & supervised\\
  SVM & 74 & 6 & 0 & supervised\\
  Decision Tree & 71 & 8 & \textbf{4}& supervised \\
  Random Forest & 83 & 6 & 2 & supervised\\
  Neural Network & 83 & 8 & 2 & supervised\\
  K-means& \textbf{84} & 4 & 0 & supervised\\
  \textbf{ScarceGAN}& 81 & \textbf{28} & 2 & semi-supervised \\
  \hline
\end{tabular}
 \caption{KDDCUP99 - Rare class (binary) identification} 
 \vspace{-0.2in}
 \label{fig-kdd-rare}
\end{table}

\subsubsection{\textbf{Imbalanced class identification: }} Here, we attempt the problem of identifying `normal' network requests from the anomalous ones. Since the data contains only 20\% of the normal requests, in research so far, the intent of predicting normal from the others has been the objective. Table \ref{kdd_imbalance} lists the state of art results (discussed in \cite{icdm,anogan,pmlr}) using the various approaches. ScarceGAN outperforms on recall from all of the state of the art including AnoGAN and ALAD. ALAD \cite{icdm} outperforms on the model precision and F1-score. ScarceGAN cost functions focus specifically on recall of the positive class hence we see a lower precision. 
\begin{table}[t]
  \centering
  \footnotesize
 \begin{tabular}{||c c c c c||}
 \hline
 \multirow{2}*{Model} & &  Performance&  & \multirow{2}*{Approach} \\
                   & Precision & Recall & F1 &\\
 \hline  
  DSEBM-r \cite{pmlr} & 0.8521 & 0.6472 & 0.7328 & supervised\\
  DSEBM-e \cite{pmlr} & 0.8619 & 0.6446 & 0.7399 & supervised\\
  ALAD \cite{icdm} & \textbf{0.94} & 0.957 & \textbf{0.9501} & unsupervised\\
  AnoGAN \cite{icdm} &  0.878 & 0.829 & 0.8865 & unsupervised\\
  \textbf{ScarceGAN}& 0.73 & \textbf{0.99} & 0.85 & semi-supervised \\
  \hline
\end{tabular}
 \caption{KDDCUP99 - imbalance identification} 
 \label{kdd_imbalance}
\vspace{-0.3in}
\end{table}
We conclude that ScarceGAN outperforms on the present state of the art results for rare class identification (KDDCUP99). It outperforms the recall benchmarks established ALAD paper \cite{icdm} for the positive imbalanced class identification. It is also worth to mention:
\emph{ScarceGAN leveraged only a small supervised prior, specifically prior knowledge on positive samples to significantly boost its recall.}

\vspace{-0.1in}

\section{Conclusion}
\label{sec:conclusion}
We introduce ScarceGAN which focuses on identification of scarce samples from multi-dimensional longitudinal telemetry data with small and weak label prior. By leveraging a `leeway' class, the discriminator gets relaxed from trying to over fit on noisy negative prior, thereby improving the quality of its negative class predictions. By differentiating the loss objectives of the positive and negative samples, it maintains a high recall of over 85\% on the scarce class with very minimal verbosity, for identification of risky skill game players. Additionally, ScarceGAN demonstrates its wide spread applicability in identification of rare classes with as much as 0.09\% imbalance over existing SOTA approaches.  We establish ScarceGAN to be one of new competitive benchmark frameworks in the rare class identification for longitudinal telemetry data. 
\balance

\begin{thebibliography}{60}


\ifx \showCODEN    \undefined \def \showCODEN     #1{\unskip}     \fi
\ifx \showDOI      \undefined \def \showDOI       #1{#1}\fi
\ifx \showISBNx    \undefined \def \showISBNx     #1{\unskip}     \fi
\ifx \showISBNxiii \undefined \def \showISBNxiii  #1{\unskip}     \fi
\ifx \showISSN     \undefined \def \showISSN      #1{\unskip}     \fi
\ifx \showLCCN     \undefined \def \showLCCN      #1{\unskip}     \fi
\ifx \shownote     \undefined \def \shownote      #1{#1}          \fi
\ifx \showarticletitle \undefined \def \showarticletitle #1{#1}   \fi
\ifx \showURL      \undefined \def \showURL       {\relax}        \fi
\providecommand\bibfield[2]{#2}
\providecommand\bibinfo[2]{#2}
\providecommand\natexlab[1]{#1}
\providecommand\showeprint[2][]{arXiv:#2}

\bibitem[Fac(2004)]%
        {Facebook}
 \bibinfo{year}{2004}\natexlab{}.
\newblock \bibinfo{booktitle}{\emph{Facebook, Inc.}}
\newblock
\urldef\tempurl%
\url{www.facebook.com}
\showURL{%
\tempurl}


\bibitem[pml(2016)]%
        {pmlr}
 \bibinfo{year}{2016}\natexlab{}.
\newblock \showarticletitle{Deep Structured Energy Based Models for Anomaly Detection}. In \bibinfo{booktitle}{\emph{Proceedings of The 33rd International Conference on Machine Learning}} \emph{(\bibinfo{series}{Proceedings of Machine Learning Research})}. \bibinfo{publisher}{PMLR}.
\newblock


\bibitem[gam(2019)]%
        {gaming}
 \bibinfo{year}{2019}\natexlab{}.
\newblock \showarticletitle{Cognitive abilities, non-cognitive skills, and gambling behaviors}.
\newblock \bibinfo{journal}{\emph{Journal of Economic Behavior \& Organization}} (\bibinfo{year}{2019}).
\newblock


\bibitem[Arjovsky et~al\mbox{.}(2017)]%
        {wasserstein}
\bibfield{author}{\bibinfo{person}{Martin Arjovsky}, \bibinfo{person}{Soumith Chintala}, {and} \bibinfo{person}{Léon Bottou}.} \bibinfo{year}{2017}\natexlab{}.
\newblock \bibinfo{title}{Wasserstein GAN}.
\newblock
\newblock
\showeprint[arxiv]{1701.07875}~[stat.ML]


\bibitem[Asokan and Seelamantula(2020)]%
        {rumigan}
\bibfield{author}{\bibinfo{person}{Siddarth Asokan} {and} \bibinfo{person}{Chandra Seelamantula}.} \bibinfo{year}{2020}\natexlab{}.
\newblock \showarticletitle{Teaching a GAN What Not to Learn}. In \bibinfo{booktitle}{\emph{Advances in Neural Information Processing Systems}}.
\newblock


\bibitem[Beggel et~al\mbox{.}(2020)]%
        {anomaly-aae}
\bibfield{author}{\bibinfo{person}{Laura Beggel}, \bibinfo{person}{Michael Pfeiffer}, {and} \bibinfo{person}{Bernd Bischl}.} \bibinfo{year}{2020}\natexlab{}.
\newblock \bibinfo{booktitle}{\emph{Robust Anomaly Detection in Images Using Adversarial Autoencoders}}.
\newblock \bibinfo{pages}{206--222}.
\newblock
\showISBNx{978-3-030-46149-2}
\urldef\tempurl%
\url{https://doi.org/10.1007/978-3-030-46150-8_13}
\showDOI{\tempurl}


\bibitem[Bengio et~al\mbox{.}(2014)]%
        {bengio}
\bibfield{author}{\bibinfo{person}{Yoshua Bengio}, \bibinfo{person}{\'{E}~ric Thibodeau-Laufer}, \bibinfo{person}{Guillaume Alain}, {and} \bibinfo{person}{Jason Yosinski}.} \bibinfo{year}{2014}\natexlab{}.
\newblock \showarticletitle{Deep Generative Stochastic Networks Trainable by Backprop}. In \bibinfo{booktitle}{\emph{Proceedings of the 31st International Conference on International Conference on Machine Learning - Volume 32}} \emph{(\bibinfo{series}{ICML'14})}.
\newblock


\bibitem[Braverman and Shaffer(2010)]%
        {gambling1}
\bibfield{author}{\bibinfo{person}{Julia Braverman} {and} \bibinfo{person}{Howard~J. Shaffer}.} \bibinfo{year}{2010}\natexlab{}.
\newblock \showarticletitle{{How do gamblers start gambling: identifying behavioural markers for high-risk internet gambling}}.
\newblock \bibinfo{journal}{\emph{European Journal of Public Health}} (\bibinfo{year}{2010}).
\newblock
\urldef\tempurl%
\url{https://doi.org/10.1093/eurpub/ckp232}
\showURL{%
\tempurl}


\bibitem[Cemiloglu et~al\mbox{.}(2020)]%
        {gambling2}
\bibfield{author}{\bibinfo{person}{Deniz Cemiloglu}, \bibinfo{person}{Emily Arden-Close}, \bibinfo{person}{Sarah Hodge}, \bibinfo{person}{Theodoros Kostoulas}, \bibinfo{person}{Maris Catania}, {and} \bibinfo{person}{Raian Ali}.} \bibinfo{year}{2020}\natexlab{}.
\newblock \showarticletitle{Towards Ethical Requirements for Addictive Technology: The Case of Online Gambling}.
\newblock
\urldef\tempurl%
\url{https://doi.org/10.1109/REthics51204.2020.00007}
\showDOI{\tempurl}


\bibitem[Chawla et~al\mbox{.}(2002)]%
        {smote}
\bibfield{author}{\bibinfo{person}{Nitesh~V. Chawla}, \bibinfo{person}{Kevin~W. Bowyer}, \bibinfo{person}{Lawrence~O. Hall}, {and} \bibinfo{person}{W.~Philip Kegelmeyer}.} \bibinfo{year}{2002}\natexlab{}.
\newblock \showarticletitle{SMOTE: Synthetic Minority over-Sampling Technique}.
\newblock \bibinfo{journal}{\emph{J. Artif. Int. Res.}} (\bibinfo{date}{June} \bibinfo{year}{2002}), \bibinfo{pages}{321–357}.
\newblock


\bibitem[Chunkai et~al\mbox{.}(2018)]%
        {vae-oversampling}
\bibfield{author}{\bibinfo{person}{Zhang Chunkai}, \bibinfo{person}{Zhou Ying}, \bibinfo{person}{Yingyang Chen}, \bibinfo{person}{Yepeng Deng}, \bibinfo{person}{Xuan Wang}, \bibinfo{person}{Lifeng Dong}, {and} \bibinfo{person}{Haoyu Wei}.} \bibinfo{year}{2018}\natexlab{}.
\newblock \bibinfo{booktitle}{\emph{Over-Sampling Algorithm Based on VAE in Imbalanced Classification}}.
\newblock \bibinfo{publisher}{Cloud Computing - CLOUD 2018}.
\newblock


\bibitem[Code(2021)]%
        {github}
\bibfield{author}{\bibinfo{person}{ScarceGAN~Source Code}.} \bibinfo{year}{2021}\natexlab{}.
\newblock \bibinfo{booktitle}{\emph{ScarceGAN: Discriminative Classification Framework for Rare Class Identification for Longitudinal Data with Weak Prior}}.
\newblock
\urldef\tempurl%
\url{https://github.com/scarce-user-53/ScarceGAN}
\showURL{%
\tempurl}


\bibitem[Dai et~al\mbox{.}(2017)]%
        {badgan}
\bibfield{author}{\bibinfo{person}{Zihang Dai}, \bibinfo{person}{Zhilin Yang}, \bibinfo{person}{Fan Yang}, \bibinfo{person}{William~W Cohen}, {and} \bibinfo{person}{Russ~R Salakhutdinov}.} \bibinfo{year}{2017}\natexlab{}.
\newblock \showarticletitle{Good Semi-supervised Learning That Requires a Bad GAN}. In \bibinfo{booktitle}{\emph{Advances in Neural Information Processing Systems}}.
\newblock


\bibitem[Dennis et~al\mbox{.}(2019)]%
        {nips_lstm}
\bibfield{author}{\bibinfo{person}{Don Dennis}, \bibinfo{person}{Durmus Alp~Emre Acar}, \bibinfo{person}{Vikram Mandikal}, \bibinfo{person}{Vinu~Sankar Sadasivan}, \bibinfo{person}{Venkatesh Saligrama}, \bibinfo{person}{Harsha~Vardhan Simhadri}, {and} \bibinfo{person}{Prateek Jain}.} \bibinfo{year}{2019}\natexlab{}.
\newblock \showarticletitle{Shallow RNN: Accurate Time-series Classification on Resource Constrained Devices}. In \bibinfo{booktitle}{\emph{Advances in Neural Information Processing Systems}}. \bibinfo{publisher}{Curran Associates, Inc.}
\newblock


\bibitem[Divekar et~al\mbox{.}(2018)]%
        {intrusion2}
\bibfield{author}{\bibinfo{person}{Abhishek Divekar}, \bibinfo{person}{Meet Parekh}, \bibinfo{person}{Vaibhav Savla}, \bibinfo{person}{Rudra Mishra}, {and} \bibinfo{person}{Mahesh Shirole}.} \bibinfo{year}{2018}\natexlab{}.
\newblock \showarticletitle{Benchmarking datasets for Anomaly-based Network Intrusion Detection: KDD CUP 99 alternatives}. In \bibinfo{booktitle}{\emph{2018 IEEE 3rd International Conference on Computing, Communication and Security (ICCCS)}}. \bibinfo{pages}{1--8}.
\newblock
\urldef\tempurl%
\url{https://doi.org/10.1109/CCCS.2018.8586840}
\showDOI{\tempurl}


\bibitem[Documentation(2017a)]%
        {shapelet}
\bibfield{author}{\bibinfo{person}{Python Documentation}.} \bibinfo{year}{2017}\natexlab{a}.
\newblock \bibinfo{booktitle}{\emph{A Python Package for Time Series Classification}}.
\newblock
\urldef\tempurl%
\url{https://pyts.readthedocs.io/en/stable/auto_examples/transformation/plot_shapelet_transform.html}
\showURL{%
\tempurl}


\bibitem[Documentation(2017b)]%
        {weasel}
\bibfield{author}{\bibinfo{person}{Python Documentation}.} \bibinfo{year}{2017}\natexlab{b}.
\newblock \bibinfo{booktitle}{\emph{WEASEL + MUSE}}.
\newblock
\urldef\tempurl%
\url{https://pyts.readthedocs.io/en/stable/auto_examples/multivariate/plot_weasel_muse.html}
\showURL{%
\tempurl}


\bibitem[Du~Plessis et~al\mbox{.}(2015)]%
        {duplessis}
\bibfield{author}{\bibinfo{person}{Marthinus~Christoffel Du~Plessis}, \bibinfo{person}{Gang Niu}, {and} \bibinfo{person}{Masashi Sugiyama}.} \bibinfo{year}{2015}\natexlab{}.
\newblock \showarticletitle{Convex Formulation for Learning from Positive and Unlabeled Data} \emph{(\bibinfo{series}{ICML'15})}.
\newblock


\bibitem[G.(2009)]%
        {dtw}
\bibfield{author}{\bibinfo{person}{Toni G.}} \bibinfo{year}{2009}\natexlab{}.
\newblock \showarticletitle{Computing and Visualizing Dynamic Time Warping Alignments in {R}: The {dtw} Package}.
\newblock \bibinfo{journal}{\emph{Journal of Statistical Software}} (\bibinfo{year}{2009}).
\newblock


\bibitem[Goodfellow(2017)]%
        {iantutorial}
\bibfield{author}{\bibinfo{person}{Ian Goodfellow}.} \bibinfo{year}{2017}\natexlab{}.
\newblock \bibinfo{title}{NIPS 2016 Tutorial: Generative Adversarial Networks}.
\newblock
\newblock
\showeprint[arxiv]{1701.00160}~[cs.LG]


\bibitem[Goodfellow et~al\mbox{.}(2014)]%
        {gan}
\bibfield{author}{\bibinfo{person}{Ian~J. Goodfellow}, \bibinfo{person}{Jean Pouget-Abadie}, \bibinfo{person}{Mehdi Mirza}, \bibinfo{person}{Bing Xu}, \bibinfo{person}{David Warde-Farley}, \bibinfo{person}{Sherjil Ozair}, \bibinfo{person}{Aaron Courville}, {and} \bibinfo{person}{Yoshua Bengio}.} \bibinfo{year}{2014}\natexlab{}.
\newblock \showarticletitle{Generative Adversarial Nets}. In \bibinfo{booktitle}{\emph{Proceedings of the 27th International Conference on Neural Information Processing Systems - Volume 2}} \emph{(\bibinfo{series}{NIPS'14})}. \bibinfo{pages}{2672–2680}.
\newblock


\bibitem[He and Garcia(2009)]%
        {occ3}
\bibfield{author}{\bibinfo{person}{Haibo He} {and} \bibinfo{person}{E.A. Garcia}.} \bibinfo{year}{2009}\natexlab{}.
\newblock \showarticletitle{Learning from Imbalanced Data}.
\newblock \bibinfo{journal}{\emph{Knowledge and Data Engineering, IEEE Transactions on}} (\bibinfo{year}{2009}).
\newblock


\bibitem[Hinton and Salakhutdinov(2006)]%
        {Hinton504}
\bibfield{author}{\bibinfo{person}{G.~E. Hinton} {and} \bibinfo{person}{R.~R. Salakhutdinov}.} \bibinfo{year}{2006}\natexlab{}.
\newblock \showarticletitle{Reducing the Dimensionality of Data with Neural Networks}.
\newblock  (\bibinfo{year}{2006}).
\newblock
\urldef\tempurl%
\url{https://science.sciencemag.org/content/313/5786/504}
\showURL{%
\tempurl}


\bibitem[Ho and Xie(1998)]%
        {arima}
\bibfield{author}{\bibinfo{person}{S.~L. Ho} {and} \bibinfo{person}{M. Xie}.} \bibinfo{year}{1998}\natexlab{}.
\newblock \showarticletitle{The Use of ARIMA Models for Reliability Forecasting and Analysis}.
\newblock \bibinfo{journal}{\emph{Comput. Ind. Eng.}} (\bibinfo{year}{1998}).
\newblock


\bibitem[Hou et~al\mbox{.}(2018)]%
        {genpu}
\bibfield{author}{\bibinfo{person}{Ming Hou}, \bibinfo{person}{Brahim Chaib-draa}, \bibinfo{person}{Chao Li}, {and} \bibinfo{person}{Qibin Zhao}.} \bibinfo{year}{2018}\natexlab{}.
\newblock \showarticletitle{Generative Adversarial Positive-Unlabelled Learning}. In \bibinfo{booktitle}{\emph{Proceedings of the Twenty-Seventh International Joint Conference on Artificial Intelligence, {IJCAI-18}}}. \bibinfo{publisher}{International Joint Conferences on Artificial Intelligence Organization}.
\newblock


\bibitem[kaggle(2018)]%
        {kaggle}
\bibfield{author}{\bibinfo{person}{kaggle}.} \bibinfo{year}{2018}\natexlab{}.
\newblock \bibinfo{booktitle}{\emph{credit card fraud}}.
\newblock
\urldef\tempurl%
\url{https://www.kaggle.com/mlg-ulb/creditcardfraud}
\showURL{%
\tempurl}


\bibitem[Kaneko et~al\mbox{.}(2019)]%
        {noisylabel}
\bibfield{author}{\bibinfo{person}{Takuhiro Kaneko}, \bibinfo{person}{Yoshitaka Ushiku}, {and} \bibinfo{person}{Tatsuya Harada}.} \bibinfo{year}{2019}\natexlab{}.
\newblock \showarticletitle{Label-Noise Robust Generative Adversarial Networks}. In \bibinfo{booktitle}{\emph{2019 IEEE/CVF Conference on Computer Vision and Pattern Recognition (CVPR)}}.
\newblock


\bibitem[Karim et~al\mbox{.}(2017)]%
        {lstm2}
\bibfield{author}{\bibinfo{person}{Fazle Karim}, \bibinfo{person}{Somshubra Majumdar}, \bibinfo{person}{Houshang Darabi}, {and} \bibinfo{person}{Shun Chen}.} \bibinfo{year}{2017}\natexlab{}.
\newblock \showarticletitle{LSTM Fully Convolutional Networks for Time Series Classification}.
\newblock \bibinfo{journal}{\emph{IEEE Access}} (\bibinfo{date}{09} \bibinfo{year}{2017}).
\newblock
\urldef\tempurl%
\url{https://doi.org/10.1109/ACCESS.2017.2779939}
\showDOI{\tempurl}


\bibitem[kddcup99(1999)]%
        {kddcup}
\bibfield{author}{\bibinfo{person}{kddcup99}.} \bibinfo{year}{1999}\natexlab{}.
\newblock \bibinfo{booktitle}{\emph{Kddcup99}}.
\newblock
\urldef\tempurl%
\url{https://datahub.io/machine-learning/kddcup99}
\showURL{%
\tempurl}


\bibitem[Kingma et~al\mbox{.}(2014)]%
        {kingma2014}
\bibfield{author}{\bibinfo{person}{Diederik~P. Kingma}, \bibinfo{person}{Danilo~J. Rezende}, \bibinfo{person}{Shakir Mohamed}, {and} \bibinfo{person}{Max Welling}.} \bibinfo{year}{2014}\natexlab{}.
\newblock \bibinfo{title}{Semi-Supervised Learning with Deep Generative Models}.
\newblock
\newblock
\showeprint[arxiv]{1406.5298}~[cs.LG]


\bibitem[Kiryo et~al\mbox{.}(2017)]%
        {kiryo2017}
\bibfield{author}{\bibinfo{person}{Ryuichi Kiryo}, \bibinfo{person}{Gang Niu}, \bibinfo{person}{Marthinus~C. du Plessis}, {and} \bibinfo{person}{Masashi Sugiyama}.} \bibinfo{year}{2017}\natexlab{}.
\newblock \showarticletitle{Positive-Unlabeled Learning with Non-Negative Risk Estimator} \emph{(\bibinfo{series}{NIPS'17})}.
\newblock


\bibitem[Krause et~al\mbox{.}(2010)]%
        {NIPS2010}
\bibfield{author}{\bibinfo{person}{Andreas Krause}, \bibinfo{person}{Pietro Perona}, {and} \bibinfo{person}{Ryan Gomes}.} \bibinfo{year}{2010}\natexlab{}.
\newblock \showarticletitle{Discriminative Clustering by Regularized Information Maximization}. In \bibinfo{booktitle}{\emph{Advances in Neural Information Processing Systems}}.
\newblock


\bibitem[Lee and Liu(2003)]%
        {lee2003}
\bibfield{author}{\bibinfo{person}{Wee~Sun Lee} {and} \bibinfo{person}{Bing Liu}.} \bibinfo{year}{2003}\natexlab{}.
\newblock \showarticletitle{Learning with Positive and Unlabeled Examples Using Weighted Logistic Regression} \emph{(\bibinfo{series}{ICML'03})}.
\newblock


\bibitem[Li and Liu(2003)]%
        {lilearning}
\bibfield{author}{\bibinfo{person}{Xiaoli Li} {and} \bibinfo{person}{Bing Liu}.} \bibinfo{year}{2003}\natexlab{}.
\newblock \showarticletitle{Learning to classify text using positive and unlabeled data}. In \bibinfo{booktitle}{\emph{In: Proceedings of the 19th international joint conference on artificial intelligence}}. \bibinfo{pages}{587--594}.
\newblock


\bibitem[Liu et~al\mbox{.}(2002)]%
        {liu2002}
\bibfield{author}{\bibinfo{person}{Bing Liu}, \bibinfo{person}{Wee~Sun Lee}, \bibinfo{person}{Philip~S. Yu}, {and} \bibinfo{person}{Xiaoli Li}.} \bibinfo{year}{2002}\natexlab{}.
\newblock \showarticletitle{Partially Supervised Classification of Text Documents}. In \bibinfo{booktitle}{\emph{Proceedings of the Nineteenth International Conference on Machine Learning}} \emph{(\bibinfo{series}{ICML '02})}. \bibinfo{publisher}{Morgan Kaufmann Publishers Inc.}, \bibinfo{address}{San Francisco, CA, USA}, \bibinfo{pages}{387–394}.
\newblock
\showISBNx{1558608737}


\bibitem[Liu et~al\mbox{.}(2012)]%
        {occ2}
\bibfield{author}{\bibinfo{person}{Fei~Tony Liu}, \bibinfo{person}{Kai~Ming Ting}, {and} \bibinfo{person}{Zhi-Hua Zhou}.} \bibinfo{year}{2012}\natexlab{}.
\newblock \showarticletitle{Isolation-Based Anomaly Detection}.
\newblock  (\bibinfo{year}{2012}).
\newblock


\bibitem[Livera et~al\mbox{.}(2010)]%
        {tbats}
\bibfield{author}{\bibinfo{person}{Alysha Livera}, \bibinfo{person}{Rob Hyndman}, {and} \bibinfo{person}{Ralph Snyder}.} \bibinfo{year}{2010}\natexlab{}.
\newblock \showarticletitle{Forecasting Time Series With Complex Seasonal Patterns Using Exponential Smoothing}.
\newblock \bibinfo{journal}{\emph{J. Amer. Statist. Assoc.}} (\bibinfo{date}{01} \bibinfo{year}{2010}), \bibinfo{pages}{1513--1527}.
\newblock
\urldef\tempurl%
\url{https://doi.org/10.1198/jasa.2011.tm09771}
\showDOI{\tempurl}


\bibitem[Maal\o{}e et~al\mbox{.}(2016)]%
        {acgan}
\bibfield{author}{\bibinfo{person}{Lars Maal\o{}e}, \bibinfo{person}{Casper~Kaae S\o{}nderby}, \bibinfo{person}{S\o{}ren~Kaae S\o{}nderby}, {and} \bibinfo{person}{Ole Winther}.} \bibinfo{year}{2016}\natexlab{}.
\newblock \showarticletitle{Auxiliary Deep Generative Models}. In \bibinfo{booktitle}{\emph{Proceedings of the 33rd International Conference on International Conference on Machine Learning - Volume 48}} \emph{(\bibinfo{series}{ICML'16})}.
\newblock


\bibitem[Mao et~al\mbox{.}(2017)]%
        {lsgan}
\bibfield{author}{\bibinfo{person}{Xudong Mao}, \bibinfo{person}{Qing Li}, \bibinfo{person}{Haoran Xie}, \bibinfo{person}{Raymond~Y.K. Lau}, \bibinfo{person}{Zhen Wang}, {and} \bibinfo{person}{Stephen~Paul Smolley}.} \bibinfo{year}{2017}\natexlab{}.
\newblock \showarticletitle{Least Squares Generative Adversarial Networks}. In \bibinfo{booktitle}{\emph{2017 IEEE International Conference on Computer Vision (ICCV)}}.
\newblock


\bibitem[Matrix(2014)]%
        {netflix}
\bibfield{author}{\bibinfo{person}{Sidneyeve Matrix}.} \bibinfo{year}{2014}\natexlab{}.
\newblock \showarticletitle{The Netflix Effect: Teens, Binge Watching, and On-Demand Digital Media Trends}.
\newblock \bibinfo{journal}{\emph{Jeunesse: Young People, Texts, Cultures}}  \bibinfo{volume}{6} (\bibinfo{date}{01} \bibinfo{year}{2014}), \bibinfo{pages}{119--138}.
\newblock
\urldef\tempurl%
\url{https://doi.org/10.1353/jeu.2014.0002}
\showDOI{\tempurl}


\bibitem[Mirza and Osindero(2014)]%
        {cgan}
\bibfield{author}{\bibinfo{person}{Mehdi Mirza} {and} \bibinfo{person}{Simon Osindero}.} \bibinfo{year}{2014}\natexlab{}.
\newblock \bibinfo{title}{Conditional Generative Adversarial Nets}.
\newblock
\newblock


\bibitem[N~R et~al\mbox{.}(2018)]%
        {social}
\bibfield{author}{\bibinfo{person}{Ramesh N~R}, \bibinfo{person}{Pruthvi S}, {and} \bibinfo{person}{Phaneendra Mallekavu}.} \bibinfo{year}{2018}\natexlab{}.
\newblock \showarticletitle{A Comparative Study on Social Media Usage and Health Status among Students Studying in Pre-University Colleges of Urban Bengaluru}.
\newblock \bibinfo{journal}{\emph{Indian Journal of Community Medicine}}  \bibinfo{volume}{43} (\bibinfo{date}{07} \bibinfo{year}{2018}), \bibinfo{pages}{180--184}.
\newblock
\urldef\tempurl%
\url{https://doi.org/10.4103/ijcm.IJCM_285_17}
\showDOI{\tempurl}


\bibitem[Odena(2016)]%
        {odena}
\bibfield{author}{\bibinfo{person}{Augustus Odena}.} \bibinfo{year}{2016}\natexlab{}.
\newblock \bibinfo{title}{Semi-Supervised Learning with Generative Adversarial Networks}.
\newblock
\newblock
\showeprint[arxiv]{1606.01583}~[stat.ML]


\bibitem[Radford et~al\mbox{.}(2016)]%
        {representationlearning}
\bibfield{author}{\bibinfo{person}{Alec Radford}, \bibinfo{person}{Luke Metz}, {and} \bibinfo{person}{Soumith Chintala}.} \bibinfo{year}{2016}\natexlab{}.
\newblock \showarticletitle{Unsupervised Representation Learning with Deep Convolutional Generative Adversarial Networks}.
\newblock \bibinfo{journal}{\emph{CoRR}}  \bibinfo{volume}{abs/1511.06434} (\bibinfo{year}{2016}).
\newblock


\bibitem[Repository(1999)]%
        {kdd-data}
\bibfield{author}{\bibinfo{person}{Dataset Repository}.} \bibinfo{year}{1999}\natexlab{}.
\newblock \bibinfo{booktitle}{\emph{KDD CUP 99 Repository and details}}.
\newblock
\urldef\tempurl%
\url{http://kdd.ics.uci.edu/databases/kddcup99/kddcup99.html}
\showURL{%
\tempurl}


\bibitem[Rose and Dhandayudham(2014)]%
        {shopaholic}
\bibfield{author}{\bibinfo{person}{Susan Rose} {and} \bibinfo{person}{Arun Dhandayudham}.} \bibinfo{year}{2014}\natexlab{}.
\newblock \bibinfo{journal}{\emph{Journal of Behavioral Addictions JBA}} \bibinfo{volume}{3}, \bibinfo{number}{2}, \bibinfo{pages}{83 -- 89}.
\newblock
\urldef\tempurl%
\url{https://doi.org/10.1556/jba.3.2014.003}
\showDOI{\tempurl}


\bibitem[Salimans et~al\mbox{.}(2016a)]%
        {salimans}
\bibfield{author}{\bibinfo{person}{Tim Salimans}, \bibinfo{person}{Ian Goodfellow}, \bibinfo{person}{Wojciech Zaremba}, \bibinfo{person}{Vicki Cheung}, \bibinfo{person}{Alec Radford}, {and} \bibinfo{person}{Xi Chen}.} \bibinfo{year}{2016}\natexlab{a}.
\newblock \bibinfo{title}{Improved Techniques for Training GANs}.
\newblock
\newblock
\showeprint[arxiv]{1606.03498}~[cs.LG]


\bibitem[Salimans et~al\mbox{.}(2016b)]%
        {fm}
\bibfield{author}{\bibinfo{person}{Tim Salimans}, \bibinfo{person}{Ian Goodfellow}, \bibinfo{person}{Wojciech Zaremba}, \bibinfo{person}{Vicki Cheung}, \bibinfo{person}{Alec Radford}, \bibinfo{person}{Xi Chen}, {and} \bibinfo{person}{Xi Chen}.} \bibinfo{year}{2016}\natexlab{b}.
\newblock \showarticletitle{Improved Techniques for Training GANs}. In \bibinfo{booktitle}{\emph{Advances in Neural Information Processing Systems}}.
\newblock


\bibitem[Schlegl et~al\mbox{.}(2017)]%
        {anogan}
\bibfield{author}{\bibinfo{person}{T. Schlegl}, \bibinfo{person}{P. Seeböck}, \bibinfo{person}{S.~M. Waldstein}, \bibinfo{person}{U. Schmidt-Erfurth}, {and} \bibinfo{person}{G. Langs}.} \bibinfo{year}{2017}\natexlab{}.
\newblock \showarticletitle{Unsupervised anomaly detection with generative adversarial networks to guide marker discovery}. In \bibinfo{booktitle}{\emph{International Conference on Information Processing in Medical Imaging}}.
\newblock


\bibitem[Sch\"{o}lkopf et~al\mbox{.}(2001)]%
        {occ1}
\bibfield{author}{\bibinfo{person}{Bernhard Sch\"{o}lkopf}, \bibinfo{person}{John~C. Platt}, \bibinfo{person}{John~C. Shawe-Taylor}, \bibinfo{person}{Alex~J. Smola}, {and} \bibinfo{person}{Robert~C. Williamson}.} \bibinfo{year}{2001}\natexlab{}.
\newblock \showarticletitle{Estimating the Support of a High-Dimensional Distribution}.
\newblock \bibinfo{journal}{\emph{Neural Comput.}} (\bibinfo{year}{2001}).
\newblock


\bibitem[Seth et~al\mbox{.}(2020)]%
        {vae_deepanshi}
\bibfield{author}{\bibinfo{person}{Deepanshi Seth}, \bibinfo{person}{Sharanya Eswaran}, \bibinfo{person}{Tridib Mukherjee}, {and} \bibinfo{person}{Mridul Sachdeva}.} \bibinfo{year}{2020}\natexlab{}.
\newblock \showarticletitle{A Deep Learning Framework for Ensuring Responsible Play in Skill-based Cash Gaming}. In \bibinfo{booktitle}{\emph{19th {IEEE} International Conference on Machine Learning and Applications, {ICMLA} 2020, Miami, FL, USA, December 14-17, 2020}}.
\newblock


\bibitem[Springenberg(2016)]%
        {catgan}
\bibfield{author}{\bibinfo{person}{Jost Springenberg}.} \bibinfo{year}{2016}\natexlab{}.
\newblock \showarticletitle{Unsupervised and Semi-supervised Learning with Categorical Generative Adversarial Networks}. In \bibinfo{booktitle}{\emph{2016 International Conference on Learning Representations (ICLR)}}.
\newblock


\bibitem[Stokes et~al\mbox{.}(2011)]%
        {intrusion-intro}
\bibfield{author}{\bibinfo{person}{Jack Stokes}, \bibinfo{person}{John Platt}, \bibinfo{person}{Joseph Kravis}, {and} \bibinfo{person}{Michael Shilman}.} \bibinfo{year}{2011}\natexlab{}.
\newblock \showarticletitle{ALADIN: Active Learning of Anomalies to Detect Intrusions, Microsoft Research}.
\newblock  (\bibinfo{date}{04} \bibinfo{year}{2011}).
\newblock


\bibitem[Suzuki et~al\mbox{.}(2019)]%
        {shapelets_anomaly}
\bibfield{author}{\bibinfo{person}{Hiroko Suzuki}, \bibinfo{person}{Ryoko Nakamura}, \bibinfo{person}{Aozora Inagaki}, \bibinfo{person}{Isamu Watanabe}, {and} \bibinfo{person}{Tomohiro Takagi}.} \bibinfo{year}{2019}\natexlab{}.
\newblock \showarticletitle{Early Detection of Problem Gambling Based on Behavioral Changes Using Shapelets}. In \bibinfo{booktitle}{\emph{IEEE/WIC/ACM International Conference on Web Intelligence}}.
\newblock


\bibitem[Taylor and Letham(2017)]%
        {prophet}
\bibfield{author}{\bibinfo{person}{Sean Taylor} {and} \bibinfo{person}{Benjamin Letham}.} \bibinfo{year}{2017}\natexlab{}.
\newblock \showarticletitle{Forecasting at Scale}.
\newblock \bibinfo{journal}{\emph{The American Statistician}} (\bibinfo{date}{09} \bibinfo{year}{2017}).
\newblock
\urldef\tempurl%
\url{https://doi.org/10.1080/00031305.2017.1380080}
\showDOI{\tempurl}


\bibitem[Vincent et~al\mbox{.}(2008)]%
        {vincent}
\bibfield{author}{\bibinfo{person}{Pascal Vincent}, \bibinfo{person}{Hugo Larochelle}, \bibinfo{person}{Yoshua Bengio}, {and} \bibinfo{person}{Pierre-Antoine Manzagol}.} \bibinfo{year}{2008}\natexlab{}.
\newblock \showarticletitle{Extracting and Composing Robust Features with Denoising Autoencoders} \emph{(\bibinfo{series}{ICML '08})}.
\newblock


\bibitem[Xu et~al\mbox{.}(2004)]%
        {xu2004}
\bibfield{author}{\bibinfo{person}{Linli Xu}, \bibinfo{person}{James Neufeld}, \bibinfo{person}{Bryce Larson}, {and} \bibinfo{person}{Dale Schuurmans}.} \bibinfo{year}{2004}\natexlab{}.
\newblock \showarticletitle{Maximum Margin Clustering} \emph{(\bibinfo{series}{NIPS'04})}.
\newblock


\bibitem[Xu et~al\mbox{.}(2020)]%
        {ccgan}
\bibfield{author}{\bibinfo{person}{Yanwu Xu}, \bibinfo{person}{Mingming Gong}, \bibinfo{person}{Junxiang Chen}, \bibinfo{person}{Tongliang Liu}, \bibinfo{person}{Kun Zhang}, {and} \bibinfo{person}{Kayhan Batmanghelich}.} \bibinfo{year}{2020}\natexlab{}.
\newblock \showarticletitle{Generative-Discriminative Complementary Learning}.
\newblock \bibinfo{journal}{\emph{Proceedings of the AAAI Conference on Artificial Intelligence}} (\bibinfo{date}{Apr.} \bibinfo{year}{2020}).
\newblock
\urldef\tempurl%
\url{https://doi.org/10.1609/aaai.v34i04.6126}
\showDOI{\tempurl}


\bibitem[Zenati et~al\mbox{.}(2018)]%
        {icdm}
\bibfield{author}{\bibinfo{person}{Houssam Zenati}, \bibinfo{person}{Manon Romain}, \bibinfo{person}{Chuan-Sheng Foo}, \bibinfo{person}{Bruno Lecouat}, {and} \bibinfo{person}{Vijay Chandrasekhar}.} \bibinfo{year}{2018}\natexlab{}.
\newblock \showarticletitle{Adversarially Learned Anomaly Detection}. In \bibinfo{booktitle}{\emph{2018 IEEE International Conference on Data Mining (ICDM)}}. \bibinfo{pages}{727--736}.
\newblock
\urldef\tempurl%
\url{https://doi.org/10.1109/ICDM.2018.00088}
\showDOI{\tempurl}


\bibitem[Zhou and Paffenroth(2017)]%
        {kdd-autoencoders}
\bibfield{author}{\bibinfo{person}{Chong Zhou} {and} \bibinfo{person}{Randy~C. Paffenroth}.} \bibinfo{year}{2017}\natexlab{}.
\newblock \showarticletitle{Anomaly Detection with Robust Deep Autoencoders}. In \bibinfo{booktitle}{\emph{Proceedings of the 23rd ACM SIGKDD International Conference on Knowledge Discovery and Data Mining}} \emph{(\bibinfo{series}{KDD '17})}.
\newblock


\end{thebibliography}

\bibliographystyle{plain}

\end{document}